\documentclass[journal]{IEEEtran}
\usepackage{graphicx}
\usepackage{subcaption}
\usepackage{multirow}
\usepackage[table]{xcolor}
\usepackage{color}
\usepackage{colortbl}
\usepackage{tabularx}
\usepackage{bm}
\usepackage{booktabs}
\usepackage{url}
\usepackage{stackengine}
\usepackage{float}
\usepackage{verbatim}
\usepackage{array}
\usepackage{algorithm}
\usepackage{algorithmic}
\usepackage[colorlinks,linkcolor=blue]{hyperref}

\begin{document}

\title{DiResNet: Direction-aware Residual Network for Road Extraction in VHR Remote Sensing Images}

\author{Lei~Ding and Lorenzo~Bruzzone,~\IEEEmembership{Fellow,~IEEE}~
\thanks{L. Ding and L. Bruzzone are with the Department
of Information Engineering and Computer Science, University of Trento,
38123 Trento, Italy (E-mail: lei.ding@unitn.it, lorenzo.bruzzone@unitn.it).}
\thanks{This work is supported by the scholarship from China Scholarship Council under the grant NO.201703170123}}

\markboth{IEEE TRANSACTIONS ON GEOSCIENCE AND REMOTE SENSING}%
{Shell \MakeLowercase{\textit{et al.}}: Bare Demo of IEEEtran.cls for IEEE Journals}

\maketitle

\begin{abstract}
The binary segmentation of roads in very high resolution (VHR) remote sensing images (RSIs) has always been a challenging task due to factors such as occlusions (caused by shadows, trees, buildings, etc.) and the intra-class variances of road surfaces. The wide use of convolutional neural networks (CNNs) has greatly improved the segmentation accuracy and made the task end-to-end trainable. However, there are still margins to improve in terms of the completeness and connectivity of the results. In this paper, we consider the specific context of road extraction and present a direction-aware residual network (DiResNet) that includes three main contributions: 1) An asymmetric residual segmentation network with deconvolutional layers and a structural supervision to enhance the learning of road topology (DiResSeg); 2) a pixel-level supervision of local directions to enhance the embedding of linear features; 3) A refinement network to optimize the segmentation results (DiResRef). Ablation studies on two benchmark datasets (the Massachusetts dataset and the DeepGlobe dataset) have confirmed the effectiveness of the presented designs. Comparative experiments with other approaches show that the proposed method has advantages in both overall accuracy and F1-score. The code is available at:
\href{https://github.com/ggsDing/DiResNet}{\textit{https://github.com/ggsDing/DiResNet}}.
\end{abstract}


\section{Introduction}\label{sc1}
Road extraction from very high resolution (VHR) remote sensing images (RSIs) is essential for the mapping and updating of geographic information systems (GIS). This task has been studied for decades but we have not satisfactory automatic solution yet. This is due to the special characteristics of roads. Compared with other compact ground objects (such as buildings and water), roads in VHR RSIs appear to be elongated regions with similar spectral and texture patterns. Additionally, roads have fixed width and limited curvatures, and they are not suddenly interrupted \cite{wang2016review}. To model these geometric features, the road extraction algorithms are expected to have a certain level of optimization and regularization of the results to reduce the discontinuities and false alarms.

Conventional expert-knowledge based methods for road extraction usually combine multiple edge detection, tracking, region clustering and filtering algorithms to obtain integrated results, since any single algorithm cannot model the complex structure of roads \cite{wang2016review}. This often makes the results parameter-dependent and leads to error accumulation problems. The rise of convolutional neural networks (CNNs) makes it possible to model roads in an end-to-end manner and generalize the results to large volumes of data. Accordingly, due to the great feature embedding power of CNNs, the accuracy of road extraction has been significantly improved. In consequence, since around 2017 the CNN-based methods have been the mainstream in road extraction \cite{cheng2017casnet}.

In the CNN-based approaches, road extraction is viewed as a binary segmentation problem. Cascaded convolutional layers are employed to model the spectral and spatial distribution of roads, followed by a classifier to densely discriminate the pixel categories (roads or non-roads). Compared to conventional methods based on hand-crafted features, if properly trained on a large number of representative annotated samples, CNNs are able to learn high-level semantic features of the roads automatically, and thus can be considered as a powerful feature extractor and classifier.

Some remaining problems in CNN-based road extraction methods are the recognition of the spectral outliers and the recovering of occluded areas (e.g. caused by shadows, trees, buildings and vehicles). These problems have been alleviated due to the encoding design of CNNs that aggregates local contextual information. However, there are still discontinuities in road segmentation maps. A possible solution to these problems is to enhance the embedding of linear features within the CNN architectures.

In this paper we address the above-mentioned problems by proposing a direction-aware residual network that adds a supervision to force the network to learn directional features. In this way, the learned network is direction-sensitive and the linear features are strengthened. Moreover, most literature works employ UNet-like architectures \cite{ronneberger2015unet}. They typically contain symmetric designs with connections to the low-level features to recover the spatial details. Although this skip connection design can provide spatial details, we argue that it has the side effect of aggravating the occlusions and fragmenting the results. By contrast, we employ ResNet as the backbone network with two additional designs: a structural loss function to enhance the learning of the road topology, and a decoder network to smoothly enlarge the feature maps. Additionally, a refinement sub-net is designed to optimize the segmentation results.

To summarize, the main contributions of this work are as follow:
\begin{enumerate}
    \item Designing a residual segmentation network (DiResSeg) with deconvolutional layers and structural supervision for the task of road extraction. This network design is aimed at enhancing the structural completeness of the road networks.
    \item Introducing a direction supervision to the network. This enables the learned model to be direction-aware, thus strengthens the detection of linear features.
    \item Introducing a refinement sub-net (DiResRef) to optimize the road extraction results.
    \item Performing ablation studies and comparative experiments on two benchmark datasets (the Massachusetts dataset and the DeepGlobe dataset) to verify the effectiveness of the introduced designs and the overall architecture (DiResNet).
\end{enumerate}

The remainder of this paper is organized as follows. Section \ref{sc2} introduces the related works on road extraction in VHR RSIs. Section \ref{sc3} illustrates the proposed method in details. Section \ref{sc4} describes the implementation details and the experimental settings. Section \ref{sc5} presents the results and analyzes the effect of the proposed method. Section \ref{sc6} draws the conclusions of this study.

\section{Related Work}\label{sc2}
\begin{figure*}[htbp]
    \centering
    \includegraphics[width=1\linewidth]{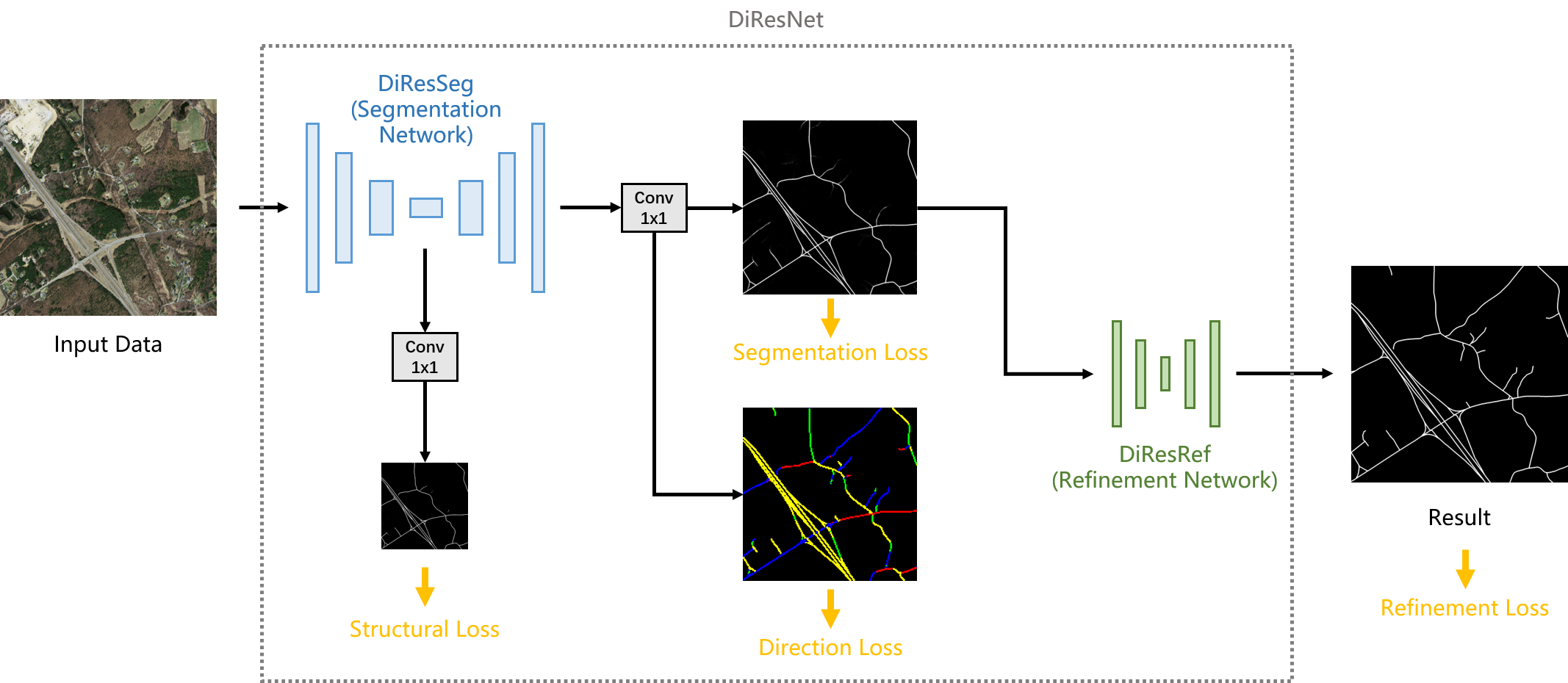}
    \caption{The proposed direction-aware residual network (DiResNet).}
    \label{FigOverview}
\end{figure*}
The literature works on automatic road extraction can be divided into two categories: expert knowledge-based methods and CNN-based ones. Although the CNN-based methods have advantages in accuracy and generalization capabilities, previous works provide inspiration on how to utilize the spatial and spectral properties of roads. In this section we briefly review these two categories of methods.
 
 \subsection{Expert Knowledge-Based Methods}
 Previous works on road extraction before the emergence of CNNs generally contain two essential steps: 1) the segmentation of roads, and 2) the refinement of the classified road segments.
 \subsubsection{Segmentation of roads}
 Literature methods on road segmentation are based on either the local spectral homogeneity or the intensity contrast of road surfaces. There are supervised methods that require training samples and unsupervised ones that operate without any labeled data. Typical supervised segmentation methods employ classifiers like the support vector machine (SVM) to label pixels based on the spectral values \cite{das2011use}. Some early works on used simple neural networks as feature extractor \cite{mnih2010learning}. However, these networks only contained a single hidden layer thus with limited capabilities to capture the problem complexity. Unsupervised segmentation methods can be further divided into edge-detection based and object-based ones. 
 
 The edge detection based methods are suitable for detecting ridge-like linear features. The Canny detector is one of the most widely used algorithms to extract road candidates \cite{trinder1998automatic}\cite{yin2015ant}\cite{sghaier2015road}. In \cite{das2011use}, the gradients obtained by Canny detectors are followed by a singular value decomposition to extract the road boundaries. In \cite{shao2010application}, 1-dimensional filtering operators are used to detect edges. The detection of local directionality is presented in \cite{zang2016road} and shows better performance compared with the Sobel operator. Hough transform is another commonly used algorithm to detect the dominant linear features in an image \cite{gamba2006improving}.
 
Meanwhile, object-based methods are applicable to the extraction of ribbon-like structures. A typical strategy is to employ clustering techniques (based on spectral and texture features) to obtain candidate super-pixels, after which applying tracking or grouping algorithms to obtain road segments \cite{yuan2011legion}. The clustering methods are usually bottom-up pixel-merging algorithms \cite{huang2009road}, some of which are implemented by the eCognition commercial software \cite{yin2015ant}. There are also plenty of works that employ angular operators to extract roads. In \cite{hu2007road}, the concept of road footprint is introduced to measure the shape of neighbourhood pixels and track the road directions. The work in \cite{ding2016road} further merges the direction-homogeneous pixels into candidate road segments.
 
 \subsubsection{Refinement of road segments}
 After the coarse segmentation of roads, candidate road-like objects are presented in the binary maps. Typically, a filtering operation is applied to these maps to remove the false alarms. Several works employ geometric calculations to discriminate the shape of candidate regions. In \cite{miao2012road}, candidate road segments are classified based on the length-width ratio of their minimum bounding rectangles. In \cite{miao2014method}, the second-order moments of segments are used to filter the non-road ones. Angular operators have also been used to measure the shape of segments in binary maps based on their circularity and rectangularity \cite{zhang2006benefit}\cite{ding2015using}.
 
 Another refinement process of the results is the optimization of the extracted road segments. This process generally includes thresholding calculations based on the geometric parameters of segments (e.g., length, distance, orientation) to simplify the road chains \cite{gamba2006improving}, merge the overlaps \cite{xiao2019method} and connect the adjacent regions and junctions \cite{das2011use}. Tensor voting is also a frequently used method to link road segments \cite{miao2012road}. It is based on a geometric analysis of the differential information of local pixels \cite{medioni2000tensor}.
 
 \subsection{CNN-based Methods}
 Although the expert knowledge-based road extraction methods can achieve satisfactory results on some RSIs, they heavily rely on the setting of the values of many parameters. In this context, the use of CNNs brings an increase in both feature representation power and generalization ability at the cost of having a huge database of annotated samples for the training of the network. Here we briefly review the literature works on CNN-based road extraction in terms of two aspects: network designs and supervisions.
 
 \subsubsection{Network designs}
 Most existing works are derived from UNet, a CNN originally designed for medical image processing \cite{ronneberger2015unet}. Since it has a symmetric encoder-decoder design and concatenation operations between the encoder and decoders, it has the ability to preserve spatial details and is suitable for processing large scale images. In \cite{yang2019road}, the convolutional units in UNet are changed to recurrent ones, which contain summation operations between the convolutional layers to better preserve spatial information. In \cite{zhang2018roadresunet}, the UNet is combined with the residual design in the ResNet \cite{he2016resnet} architecture. The resulting ResUNet shows a better performance compared with the original UNet. A similar design is introduced in \cite{xin2019road} by combining UNet and the Dense block \cite{huang2017densely}. In \cite{xu2018road}, two attention units are incorporated into the DenseUNet network to introduce skip-layer attentions at both the global and local levels. Dilated convolutions have also been used to enlarge the receptive filed of the CNN \cite{zhang2019jointnet}. There are also CNNs designed for multi-task learning. In \cite{cheng2017casnet}, two encoder-decoder CNNs are cascaded to perform the task of road segmentation and road centerline extraction, respectively. In \cite{liu2018roadnet}, two parallel branches are added after the road segmentation network to learn road edges and centerlines simultaneously. Additionally, the generative adversarial network is introduced for the segmentation of roads in \cite{zhang2019aerial}. It includes a discriminator to improve the generation of road maps.
 
 \subsubsection{Supervisions} 
 Additional supervision or the variation of loss functions can affect the learning of features. In \cite{liu2018roadnet}, a multi-scale supervision is introduced to supervise each decoding layer. It also introduces human interactions to fix the incomplete predictions. In \cite{zhang2019topology}, the topology supervision (by centerline maps) is introduced to enable the network to better deal with occlusions. To emphasize the pixels close to road regions, a weighted loss function based on the calculation of euclidean distance has been introduced in \cite{wei2017road}. While the binary cross-entropy (BCE) loss function is commonly used for binary segmentation, the structure similarity  (SSIM) loss has been adopted in \cite{he2019road} to enhance the quality of the segmentation. In \cite{batra2019improved}, a parallel branch that learns the orientation of roads is added as an auxiliary supervision to improve the connectivity of the road features.
 
 To conclude, although there are numerous works on CNN-based road extraction in VHR RSIs, most of them are simple extensions of the commonly used CNN architectures without considering the specific context of road extraction. Thus, there are still margins to improve the accuracy of road segmentation in terms of completeness and smoothness.

\section{Proposed Direction-aware Residual Network}\label{sc3}
In this section we present a direction-aware residual network (DiResNet) integrating several network designs and auxiliary supervisions. We illustrate first the network designs and then the auxiliary supervisions.

\begin{figure}[htbp]
\centering
        \subcaptionbox{}
        {\includegraphics[width=9cm]{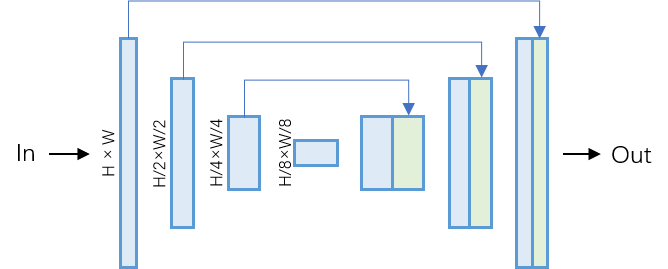}}\\
        \subcaptionbox{}
        {\includegraphics[width=4.2cm]{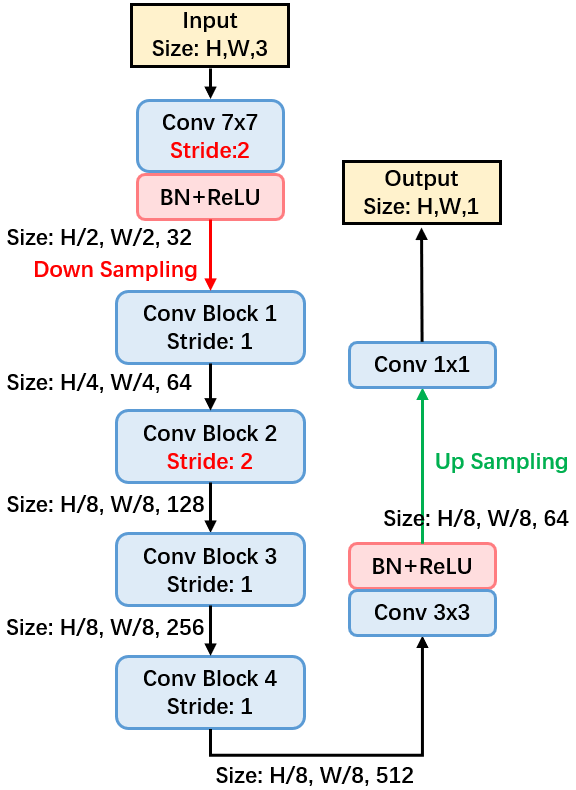}}
        \subcaptionbox{}
        {\includegraphics[width=4.4cm]{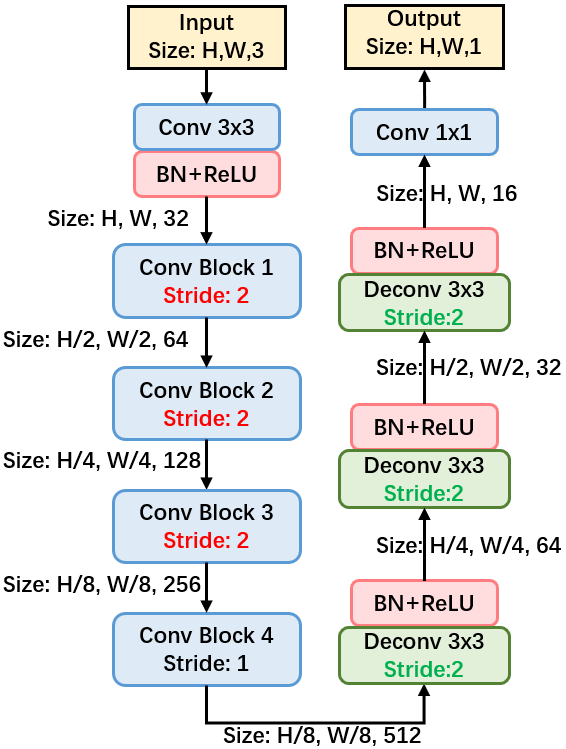}}
    \caption{Illustration of the segmentation networks. (a) Traditional UNet-like Architecture, (b) FCN, (c) DiResSeg: the designed segmentation network.}\label{FigSegNet}
\end{figure}

\subsection{Network architecture}
\subsubsection{DiResSeg: Designed Segmentation Network}

An overview of the designed direction-aware residual network (DiResNet) is shown in Fig.\ref{FigOverview}. The network consists of two sub-nets: i) a segmentation network (DiResSeg) for the coarse segmentation of roads, and ii) a refinement network (DiResRef) to optimize the segmentation results. There are also two auxiliary supervisions in the network: i) a structure supervision in the middle of the segmentation network, and ii) a direction supervision from a parallel branch of the decoder module.

Most literature works on road extraction adopt UNet-like architectures. It has been believed that the multi-scale concatenation of low-level features can improve the performance. However, recent studies found that the integration of low-level features contributes little to binary segmentation tasks, whereas it does pass through noisy information \cite{wang2018detect} and increases the computational costs \cite{wu2019cascaded}. For the task of road extraction, we expect the extracted objects to be continuous and smooth elongated regions with fixed width, while the pixel-level accurate segmentation of road boundaries is not necessary. The low-level features are generally more noisy (due to occlusions and spectral outliers). The skip connections with them may lead to uneven boundaries and interruptions. Therefore, we argue that the multi-level concatenation operations in UNet-like structures is unnecessary, if not disadvantageous, for road extraction. On the contrary, we emphasize on the embedding power of encoder networks, and present an asymmetrical encoder-decoder design with strengthened encoder and simplified decoder.

Fig.\ref{FigSegNet} shows a comparison of our network design versus the UNet-like networks. They both contain an encoder network and a decoder module. Fig.\ref{FigSegNet}(a) shows the case of a UNet-like CNN with 4 encoding layers. let us denote the encoded features as $\{E1, E2, E3, E4\}$, their corresponding spatial scaling ratios are 1, 1/2, 1/4 and 1/8, respectively. Accordingly, the decoder module contains 3 levels of upsampling and feature fusion operations. Each level of the decoded feature ${D}_{i} \in \{D_{3}, D_{2}, D_{1}\}$ is calculated as:
\begin{equation}
D_{i} =  U\{F_{U}(E_{i+1}), E_{i}\},
\end{equation}
Where $F_{U}$ denotes an upsampling or deconvolution operation and U denotes a convolution operation.

In our design in Fig.\ref{FigSegNet}(c), the encoder network is replaced to a layer-rearranged version of ResNet. Compared with the original FCN with ResNet backbone (see Fig.\ref{FigSegNet}(b)), the striding operations in our network are designed inside three convolutional blocks so that the feature size is reduced gradually. It also contains a simple decoder with 3 serial deconvolutional layers to enlarge the feature map and smooth the boundaries. In this way, the predicted road maps are closely related to the high-level features of the encoder network. Compared with UNet-like architectures, the designed segmentation network (DiResSeg) gives more focus on the completeness of the road topology, rather than the extraction of road boundaries. An auxiliary supervision is further added to improve the training of the encoder network (see \ref{sc3_supervisions} for more details).

\begin{table}[htbp]
\centering
    \caption{Comparison of model size and calculations expressed in terms of params (Mb) and FLOPs (Gbps), respectively.}
    \resizebox{1\linewidth}{!}{%
    \begin{tabular}{l|cccc}
        \toprule
        Method & UNet \cite{ronneberger2015unet} & Res-UNet \cite{zhang2018roadresunet} & DiResSeg (Resnet18) & DiResSeg (Resnet34)\\
        \hline
        Params (Mb) & 9.16 & 8.22 & 11.21 & 21.32 \\
        FLOPS (Gbps) & 221.43 & 182.21 & 62.10 & 115.31\\
        \bottomrule
    \end{tabular}}
    \label{Table.Calculations}
\end{table}

Table \ref{Table.Calculations} shows a comparison of calculation resources required by the designed DiResSeg (layer-rearranged ResNet with deconvolutional layers), UNet and ResUNet. The listed two versions of the DiResSeg use different ResNet backbones (ResNet18 and ResNet34). The floating point operations per second (FLOPS) are calculated based on an input size of [3, 320, 320]. Although the DiResSeg has a larger parameter size, its FLOPs are significantly smaller compared to the other two UNet-like CNNs. This is because the concatenation of low-level features in UNet-like structures greatly increases the computation costs, whereas it is not adopted in DiResSeg.

\subsubsection{DiResRef: Designed Refinement Network}
\begin{figure}[htbp]
\centering
        \includegraphics[width=9cm]{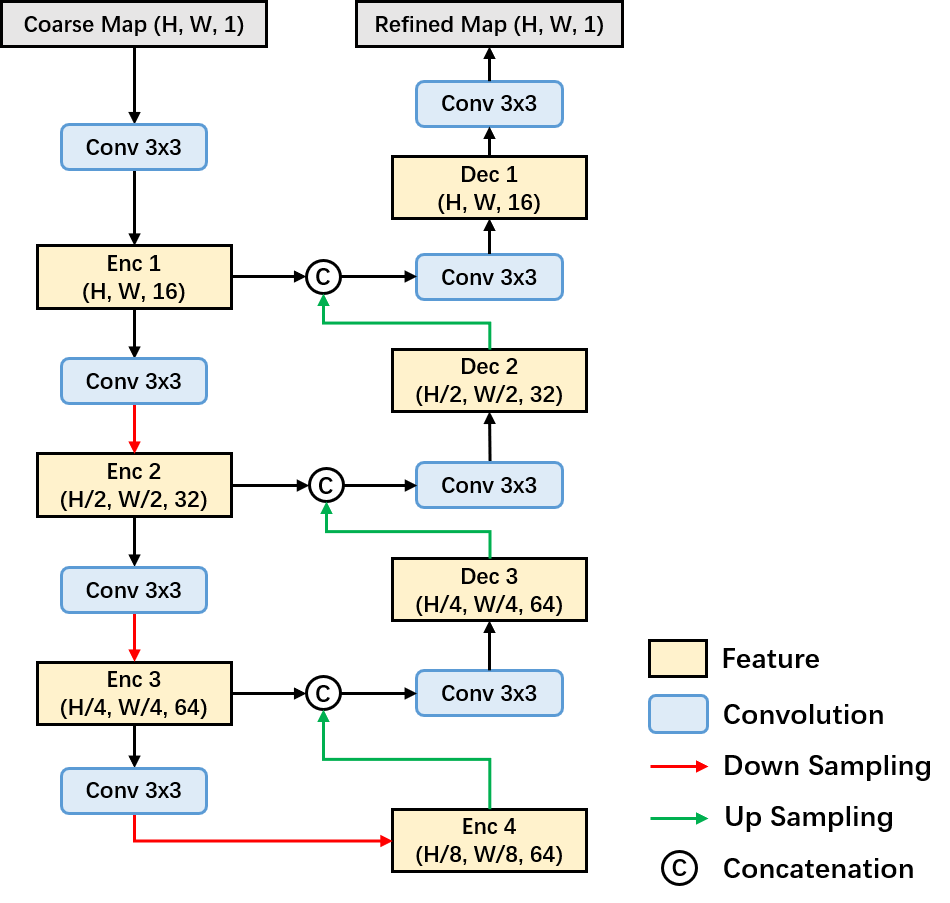}
    \caption{DiResRef: the designed refinement network.}\label{FigRefineNet}
\end{figure}

The coarsely segmented road maps may still contain interruptions and errors. An additional refinement process helps to optimize the segmentation results. An approach used in the literature work employs the tensor voting algorithm as a post-processing to the output of the CNN \cite{gao2019road}. This algorithm is able to model the underlying spatial distribution pattern of images and thus to connect the broken road segments. However, a limitation of the tensor voting algorithm is that it is based on a parameter-dependent deduction of the binary results, thus it is not stable and may produce false alarms. In this work we introduce a sub-net to perform the refinement, named as the DiResRef. It produces more stable results and makes the whole segmentation process end-to-end trainable.

The designed DiResRef is a UNet-like CNN inspired by the refine module presented in \cite{qin2019basnet}. We change the striding rate and the number of channels per-layer to adapt the network to the task of road extraction. Fig.\ref{FigRefineNet} shows the design of the DiResRef. It operates on 4 sequential encoding levels with an increasing number of channels. The input of the network is the probability map produced by the segmentation network. It produces a residual feature map which strengthens the road-like regions and suppresses the non-road ones. This network optimizes the results in various aspects, including linking the possible interruptions, removing the isolated false alarms and increasing the probability salience of the road features.

\subsection{Supervisions and Loss Functions}\label{sc3_supervisions}

The proposed direction-aware network contains different supervisions: two segmentation supervisions, a structure supervision and a direction supervision. A hybrid loss is calculated based on these supervisions, calculated as:

\begin{equation}
    loss = \alpha l_{seg} + \beta l_{struct} + \gamma l_{direct} + \theta l_{ref}
    \label{formula.loss}
\end{equation}

where $l_{struct}$, $l_{direct}$, $l_{seg}$ and $l_{ref}$ denote the losses for the structure supervision, the direction supervision, the segmentation network (DiResSeg) and the refinement network (DiResRef), respectively. $\alpha, \beta, \gamma, \theta$ are 4 weight variables for the different losses.

\subsubsection{Segmentation Supervisions}
The segmentation results are expected to be probability maps, while the reference maps are binary. Binary cross entropy loss is the most widely used function to measure the differences between predictions and targets, calculated as:
\begin{equation}
    l_{bce} = -\sum_{(r,c)}{T(r,c)log[P(r,c)]+[1-T(r,c)]log[1-P(r,c)]}
\end{equation}
where $T(r,c)\in\{0,1\}$ is the target value of pixel (r,c) and P(r,c) is the predicted probability value.

\subsubsection{Structure Supervision}
This is an auxiliary supervision added at the highest level of the encoder network, related to the down-scaled feature maps. The reference data for the structure supervision are generated by down sampling the ground truth maps (using the area interpolation). In the down-scaled maps, the width of roads is greatly reduced, thus the road boundaries are obscure. Therefore, a supervision at this level gives more attention to the center road pixels, thus strengthening the geometric structure of roads. Additionally, this auxiliary supervision is beneficial for improving the training stability of the encoder network. We deem the embedding of road structures as a regression problem and use the L1 loss to measure the structural differences:
\begin{equation}
    l_{struct} = \sum_{(r,c)}|P_s(r,c)-T_s(r,c)|
\end{equation}
where $P_s(r,c) \in [0, 1]$ and  $P_s(r,c) \in [0, 1]$ are the pixel values at the scaled target and prediction maps, respectively.

\subsubsection{Direction Supervision}

A previous study has found that learning the road orientations is beneficial to improve the connectivity of road segmentation results \cite{batra2019improved}. However, this study calculates the road orientations based on vector data, which does not apply to raster reference data. We extend this idea to common road segmentation tasks by generating the reference direction maps from the binary ground truth maps.

\begin{figure}[thpb]
\centering
    \subcaptionbox{}
    {\includegraphics[height=2.5cm]{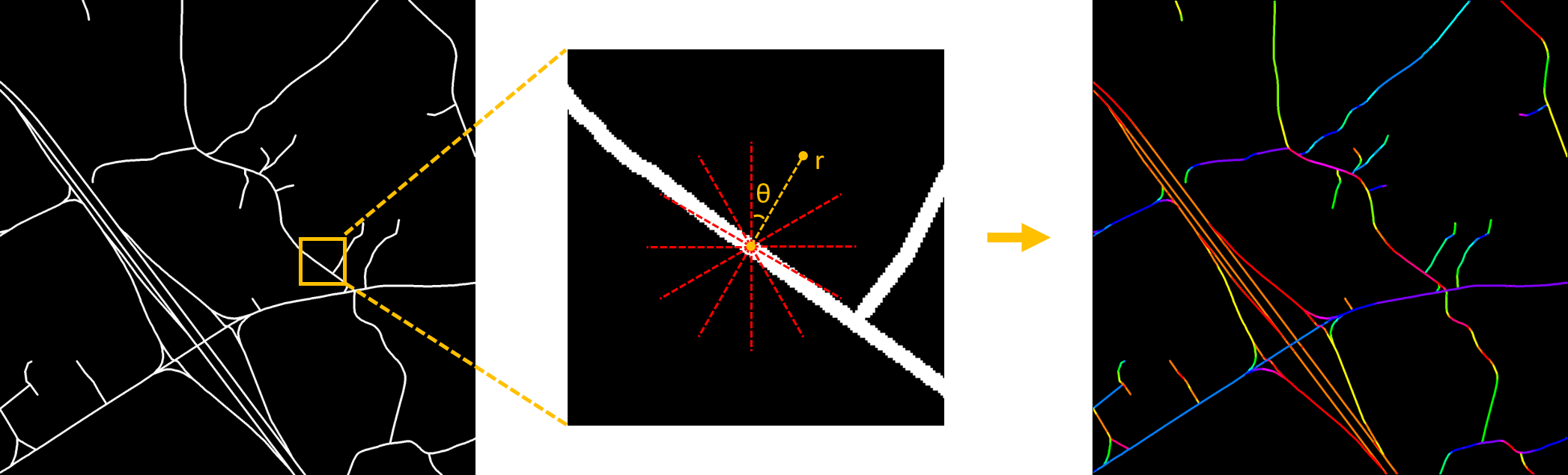}}\\
    \subcaptionbox{}
    {\includegraphics[height=2.7cm]{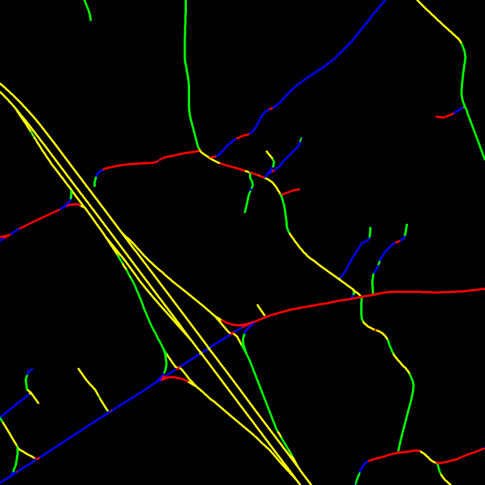}}
    \subcaptionbox{}
    {\includegraphics[height=2.7cm]{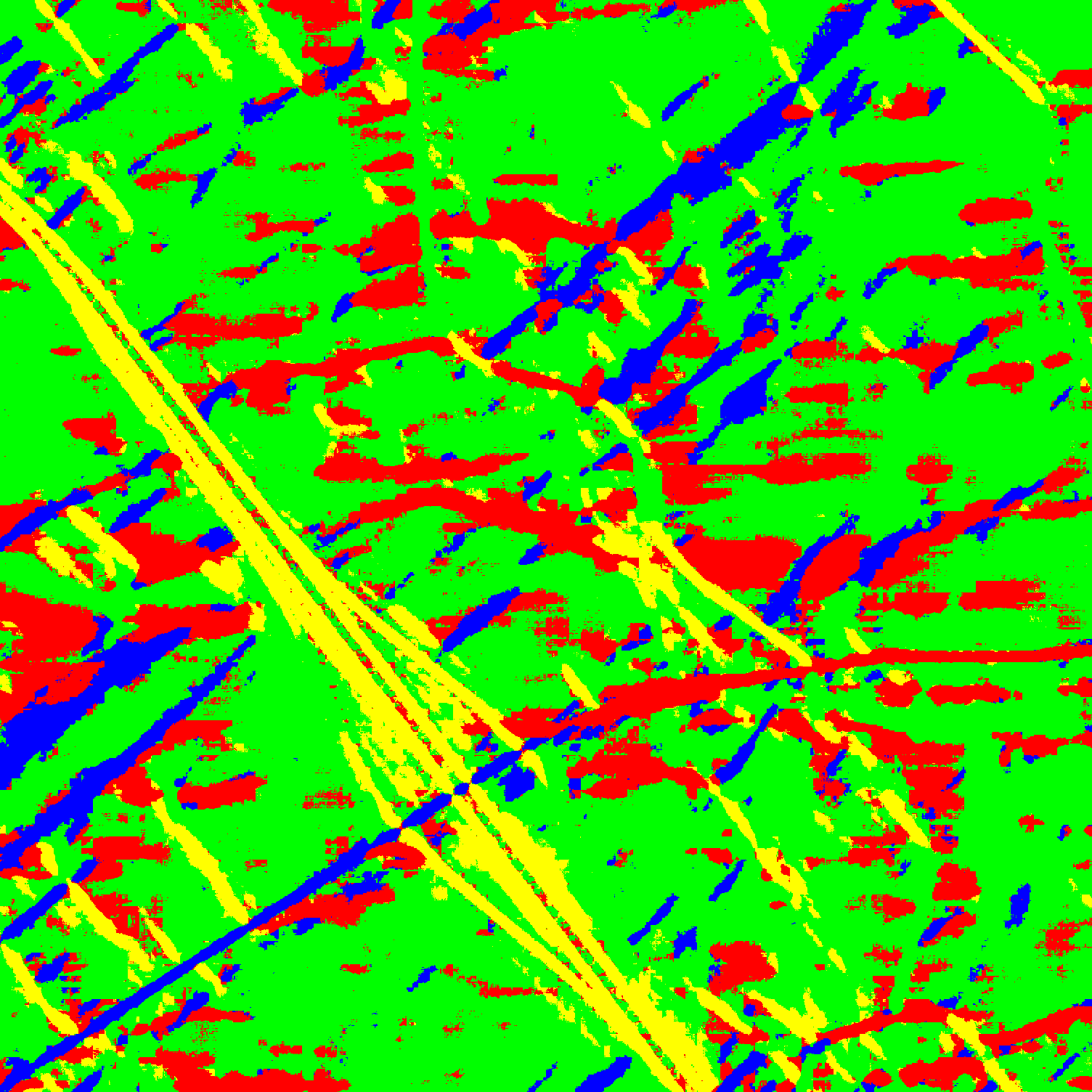}}
    \subcaptionbox{}
    {\includegraphics[height=2.7cm]{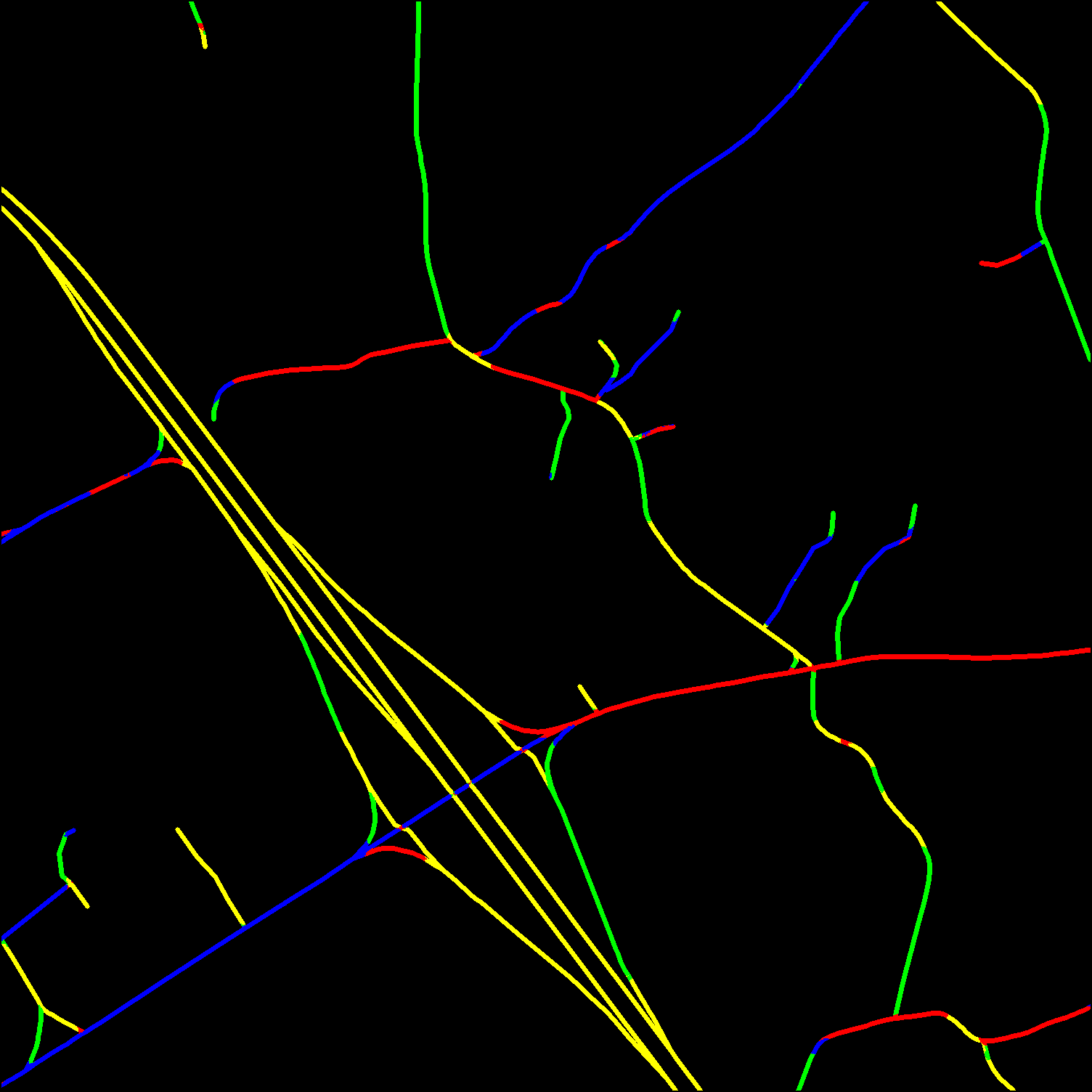}}\\
    \caption{Generation of the reference direction map. (a) Calculation of the local direction, (b) Generated reference direction map (4 main directions only), (c) Predicted direction map, (d) Predicted direction map (background pixels not shown).}
    \label{FigDMap}
\end{figure}

Fig.\ref{FigDMap} illustrates the algorithm to generate reference direction maps. The angular operators are used to measure the local direction of road pixels \cite{ding2016road}. The algorithm is as follows:

\begin{algorithm}[H]
    \caption{Algorithm for Generating the Direction Map}
    \begin{algorithmic}[1]
        \renewcommand{\algorithmicrequire}{\textbf{Input:}}
        \renewcommand{\algorithmicensure}{\textbf{Output:}}
        \REQUIRE Binary ground truth road map $T$;\\
            Parameters: detecting radius $r$, angle step $\Delta_\theta$;
        \ENSURE  Reference road direction map $T_d$;
        \FOR {$T(i,j)$ in $T$}
        \IF {($T(i,j) = 1$)}
            \FOR {$\theta$ = 0 to $\pi$ step $\Delta_\theta$}
                \STATE $d_{\theta}(i,j) = \sum_{\rho=1}^{r}T(\rho \sin \theta, \rho \cos \theta)+T(-\rho \sin \theta, -\rho \cos \theta)$
            \ENDFOR
            \STATE find $\theta_{max}$ that:\\
            \quad $d_{\theta_{max}}(i,j) = max\{d_{\theta}(i,j)\}, \theta \in [0, \pi]$
            \STATE $T_d(i,j) = \theta_{max}$
        \ELSE 
            \STATE $T_d(i,j) = invalid$
        \ENDIF
        \ENDFOR
 \RETURN $T_d$ 
    \end{algorithmic}
\end{algorithm}

The parameters $r$ and $\Delta_\theta$ are selected based on the minimum and maximum pixel width of the roads in $T$. The algorithm is implemented using convolutional layers with fixed weights, so that the reference maps can be generated dynamically during the training phase. To avoid under-fitting problems, we clip the target direction map $T_d$ to a 5 channel map representing 4 major directions and the non-road label ($T_d(r,c) \in \{0,1,2,3,4\}$). The multi-class cross entropy loss is used to evaluate the predicted direction map $P_d$:
\begin{equation}
    l_{direct} = \sum_{(r,c)} \sum_{i=1}^{N_d} \{ -P_d(r,c)[d_i] + log[ \sum_{j=1}^{N_d} exp(P_d(r,c)[j]) ] \}
\end{equation}
where $N_d$ is the number of road directions. To encourage the modeling of linear features over all areas, the non-road labels are neglected in $l_{direct}$.

This direction supervision is connected to a parallel branch of the segmentation network, thus the learned direction features can contribute to the segmentation results.

\section{Dataset Description and Design of Experiments}\label{sc4}
In this section we describe the experimental dataset, the implementation details and the evaluation metrics.

\subsection{Datasets Descriptions}
Two datasets are selected for experiments: the Massachusetts Dataset and the DeepGlobe Dataset. These are by far the two largest datasets openly available for road segmentation in VHR RSIs.

\subsubsection{Massachusetts Dataset \cite{MnihThesis}}
This is an aerial dataset collected in Massachusetts, US., covering an area of 2.25 square kilometers. The ground sampling distance (GSD) of this dataset is 1.2m per pixel. There are 1171 images in total, among which 1108 images are for training, 14 ones for validation and the remaining 49 ones for testing. Each image has 1500 $\times$ 1500 pixels. The imaged regions include urban, suburban and rural scenes. The reference maps are generated by rasterizing the vector data of road centerlines, so each road has a fixed width. 

\subsubsection{DeepGlobe Dataset \cite{demir2018deepglobe}}
This is a satellite dataset containing images collected in Thailand, Indonesia and India. It covers a total land area of 2220 $km^2$, containing both urban and suburban areas. The GSD of this dataset is 50cm per pixel. Each image has a size of 1024 $\times$ 1024 pixels. The original dataset contains 8570 images, among which 6226 training images are openly available with ground truth data. Therefore, we further divide the accessible data into the training and testing set with a ratio of 5:1. 5189 images are selected for training, the remaining 1037 images for testing. Compared to the Massachusetts dataset, more types of road surfaces are contained, thus road extraction in this dataset is more challenging.

\subsection{Implementation Details}
\begin{figure}[htbp]
\centering
        \includegraphics[width=9cm]{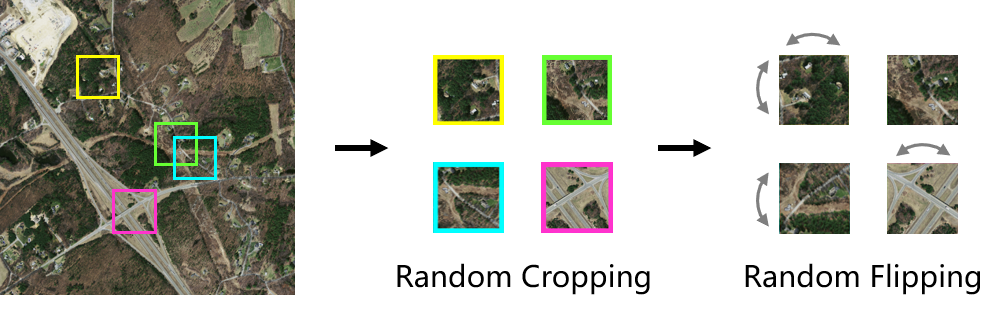}
    \caption{Illustration of the considered dynamic data augmentation process.}\label{FigAugmentation}
\end{figure}

All the experiments are performed on a workstation with 32 GB RAM and a NVIDIA Quadro P6000 GPU (23GB). The designed networks are implemented using the PyTorch library. Due to the limitation of GPU memory, the training is performed using cropped images with the spatial size of 320 $\times$ 320 pixels. To avoid the over-fitting problem, dynamic cropping and flipping operations are performed as augmentations to the dataset. The input images are first loaded and stored in the memory. During each training iteration, they are randomly cropped and flipped before being loaded to CNNs. Fig.\ref{FigAugmentation} shows our data augmentation strategy. In our implementation, each training image generates 10 cropped patches during each epoch, while the training takes 50 epochs. The training batch size is set to 16. During the validation and testing phase, full-size input images are used instead (without the cropping operation) to avoid the impact of cropping parameters. The parameters $\alpha, \beta, \gamma, \theta$ in formula \ref{formula.loss} are empirically set to 1.0, 0.5, 0.2 and 1.0, respectively. The $l_{direct}$ is assigned with a lower weight since its values are bigger.

Since the experimental datasets have different GSDs, we empirically chose different scaling rates for them in the implemented networks. The minimum down sampling rates are 1/8 and 1/16 for the Massachusetts and DeepGlobe datasets, respectively. The chosen down sampling rates are implemented in all the compared methods to ensure fairness. The ResNet18 is set as the backbone network of DiResNet and FCN for experiments on the Massachusetts Dataset, whereas ResNet34 is selected for experiments on the DeepGlobe dataset. This is because there are more types of road surfaces in the DeepGlobe dataset, which requires the encoder network to be more powerful.

\subsection{Evaluation Metrics}
We use five measurements to evaluate effectiveness of the considered methods: Precision (P), Recall (R), F1 score, overall accuracy (OA) and break-even point (BEP). These are the most widely used measurements in both road extraction and other binary segmentation tasks \cite{liu2018roadnet}. They are calculated as follow:
    \begin{equation}
        P=\frac{TP}{TP+FP}, R=\frac{TP}{TP+FN}
    \label{formular_PR}
    \end{equation}
    \begin{equation}
        F1=2\times\frac{P \times R}{P+R}, OA=\frac{TP+TN}{TP+FP+TN+FN}
    \label{formular_F1OA}
    \end{equation}
Where TP, FP, TN and FN represents true positive, false positive, true negative and false negative, respectively.
Since there is a negative correlation between the values of precision and recall (under different thresholds), we also use the break-even point as a measurement. The BEP is defined as the intersection point on the precision-recall curve, where the values of precision and recall are equal.
\section{Experimental Results}\label{sc5}
This section presents the experimental results obtained on the Massachusetts roads dataset. First an ablation study is performed to test the modules and auxiliary supervisions. Then the effects of the direction supervision and the refinement network are analyzed. Finally we compare the proposed direction-aware residual network with several literature works and analyze the results.

\subsection{Ablation Study}

\begin{figure}[thpb]
\centering
    {\includegraphics[height=0.5cm]{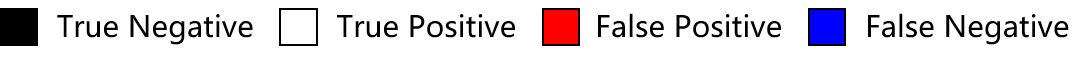}}\\
    \setlength{\tabcolsep}{1pt}
    \begin{tabular}{>{\centering\arraybackslash}m{0.5cm}>{\centering\arraybackslash}m{2cm}>{\centering\arraybackslash}m{2cm}>{\centering\arraybackslash}m{2cm}>{\centering\arraybackslash}m{2cm}}
        (a)&
        \includegraphics[width=2cm]{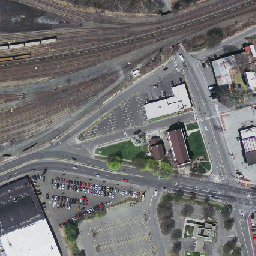} &
        \includegraphics[width=2cm]{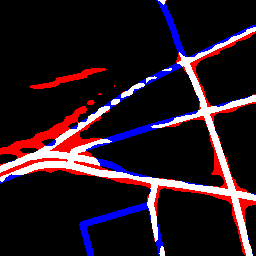} &
        \includegraphics[width=2cm]{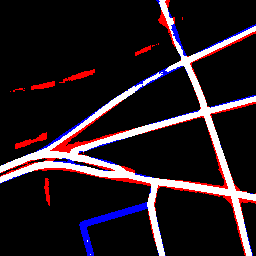} &
        \includegraphics[width=2cm]{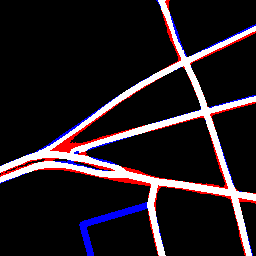}\\
        (b)&
        \includegraphics[width=2cm]{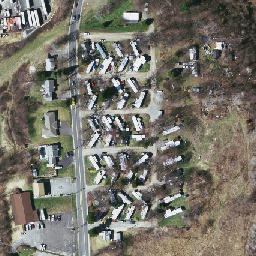} &
        \includegraphics[width=2cm]{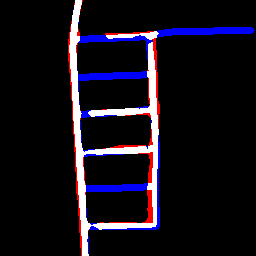} &
        \includegraphics[width=2cm]{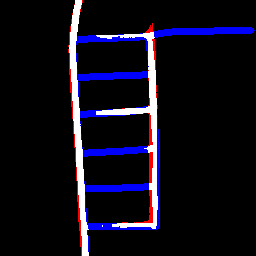} &
        \includegraphics[width=2cm]{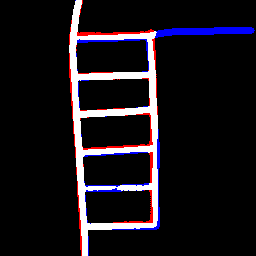}\\
        (c)&
        \includegraphics[width=2cm]{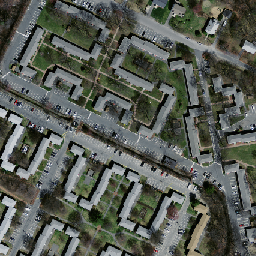} &
        \includegraphics[width=2cm]{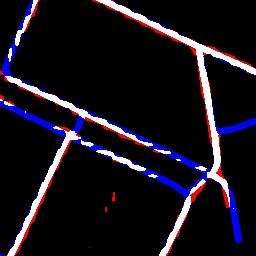} &
        \includegraphics[width=2cm]{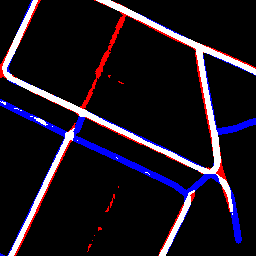} &
        \includegraphics[width=2cm]{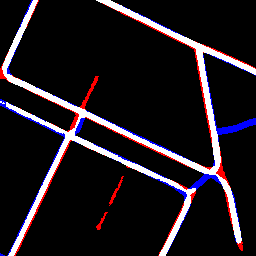}\\
        (d)&
        \includegraphics[width=2cm]{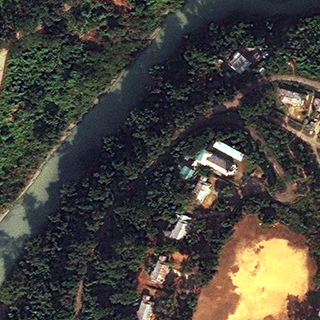} &
        \includegraphics[width=2cm]{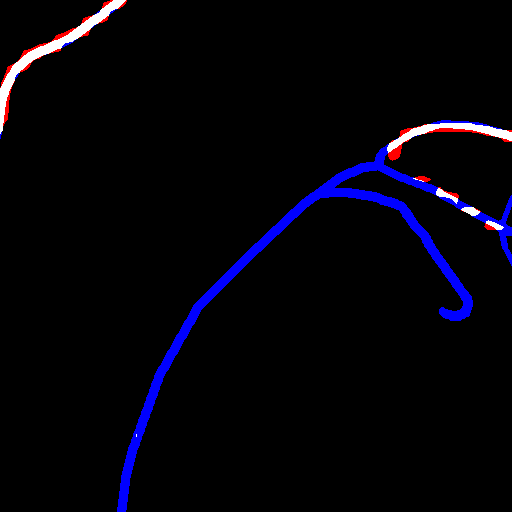} &
        \includegraphics[width=2cm]{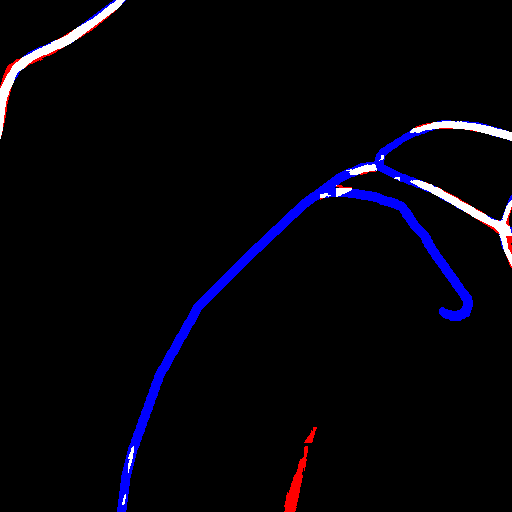} &
        \includegraphics[width=2cm]{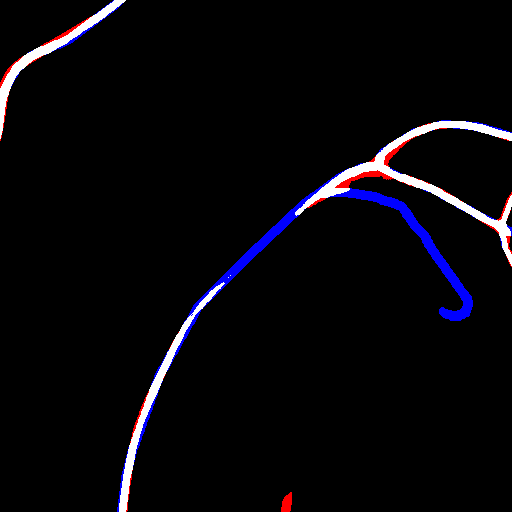}\\
        (e)&
        \includegraphics[width=2cm]{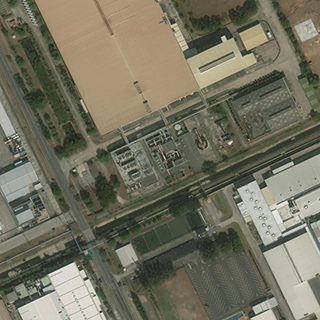} &
        \includegraphics[width=2cm]{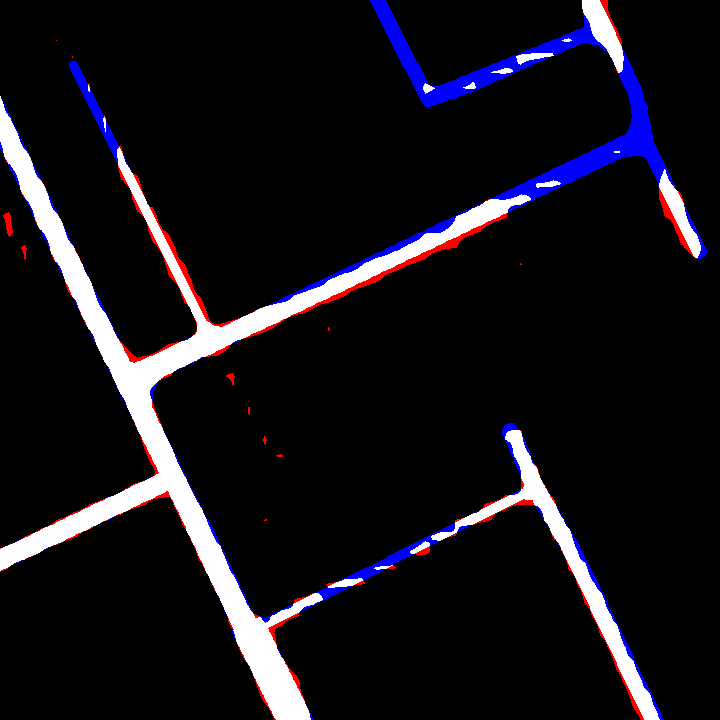} &
        \includegraphics[width=2cm]{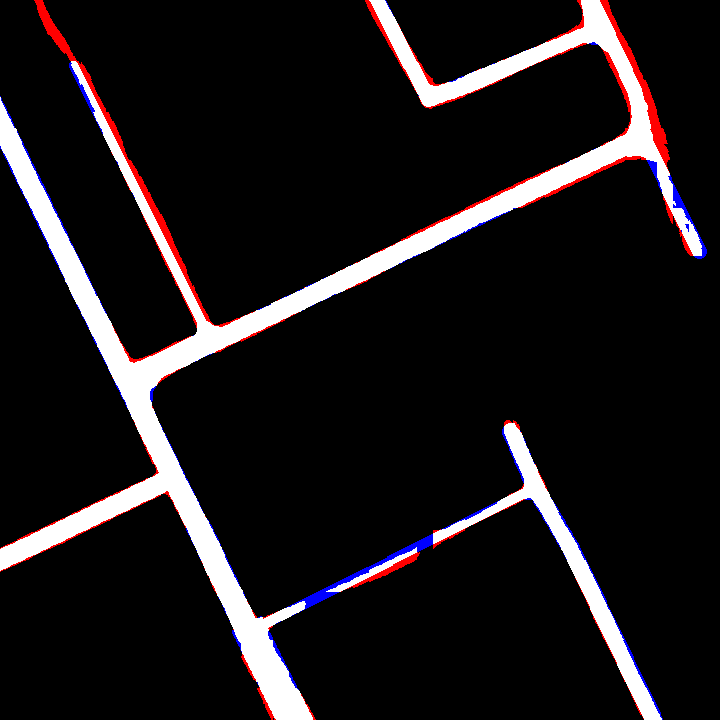} &
        \includegraphics[width=2cm]{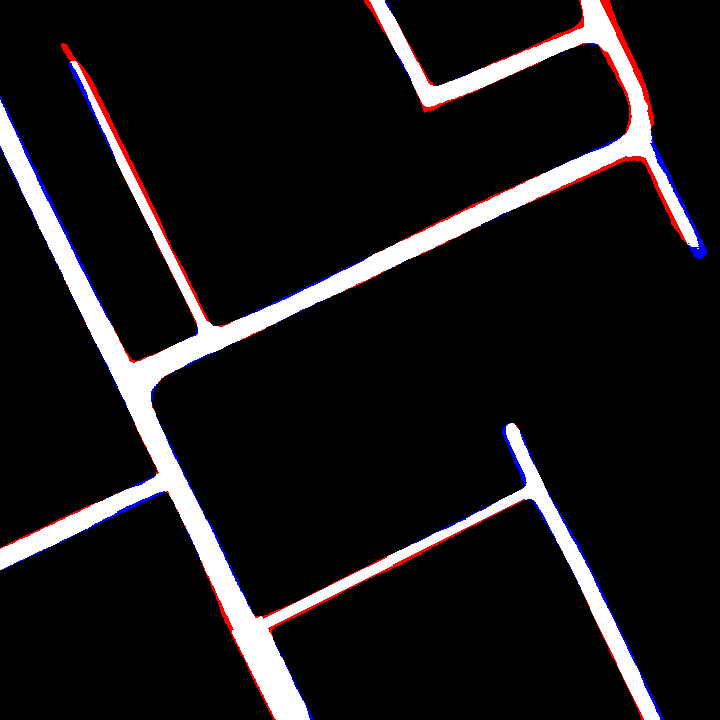}\\
        (f)&
        \includegraphics[width=2cm]{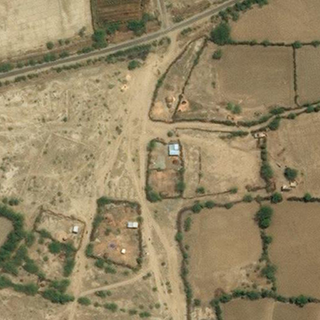} &
        \includegraphics[width=2cm]{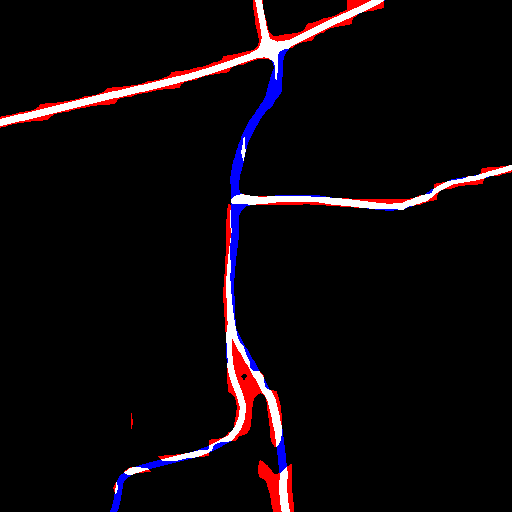} &
        \includegraphics[width=2cm]{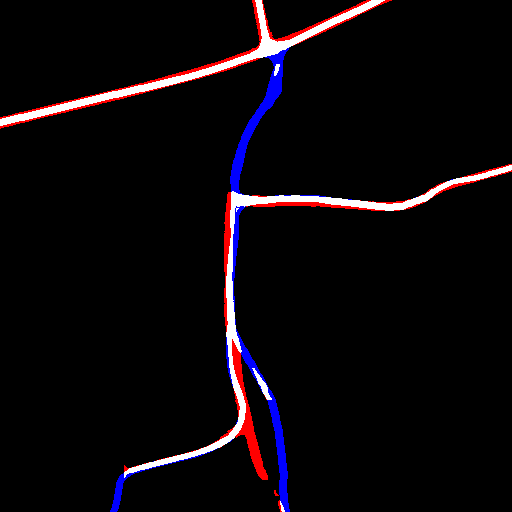} &
        \includegraphics[width=2cm]{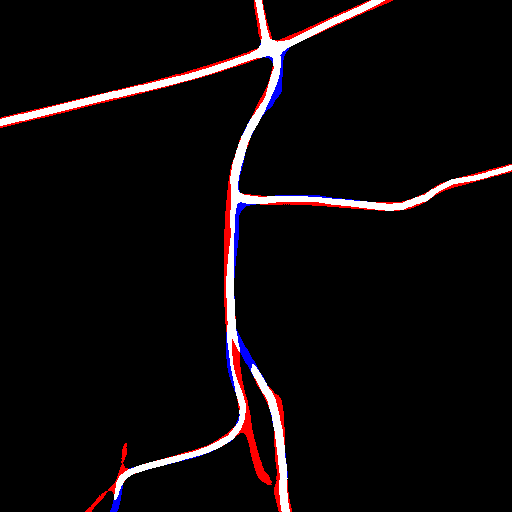}\\
        & Test image & FCN & DiResSeg & DiResNet\\
    \end{tabular}
    \caption{Example of the segmentation results (ablation study). (a)-(c) Results selected from the Massachusetts dataset, (d)-(f) Results selected from the DeepGlobe dataset.}
    \label{Fig.AblationResults}
\end{figure}

To test the effectiveness of the proposed direction-aware residual network including both the sub-nets and the auxiliary supervisions, we performed an ablation study. Since the DiResSeg is a modified version of the baseline FCN, the latter is also used in the comparison to evaluate the improvements.

\begin{table}[htbp]
    \centering
    \caption{Results of the ablation study (Massachusetts dataset).}
    \resizebox{1\linewidth}{!}{%
        \begin{tabular}{l|cc|cc|cccc}
        \toprule
            \multirow{2}*{Method} & \multicolumn{2}{c|}{Components} & \multicolumn{2}{c|}{Supervisions} & \multirow{2}*{OA(\%)}  & \multirow{2}*{P(\%)}  & \multirow{2}*{R(\%)}  & \multirow{2}*{F1(\%)} \\
            \cline{2-5}
            & DiResSeg & DiResRef & Structure & Direction \\
            \hline
            FCN &  &  &  &  & 97.70 & 74.25 & 78.93 & 76.35 \\
            \hline
            DiResSeg & $\surd$ &  &  &  & 98.05 & 78.94 & 79.86 & 79.25 \\
            DiResNet-R & $\surd$ & $\surd$ &  &  & 98.04 & 78.50 & 80.62 & 79.41  \\
            DiResNet-S & $\surd$ &  & $\surd$ &  & 98.07 & 79.44 & 79.55 & 79.33 \\
            DiResNet-D & $\surd$ & &  & $\surd$ & 98.04 & 78.57 & \textbf{80.77} & 79.48\\
            DiResNet & $\surd$ & $\surd$ & $\surd$ & $\surd$ & \textbf{98.13} & \textbf{80.38} & 79.41 & \textbf{79.70}\\
        \bottomrule
        \end{tabular}
    }
    \label{Table.AblationMas}
\end{table}

\begin{figure}[htbp]
\centering
        \subcaptionbox{}
        {\includegraphics[width=4.3cm]{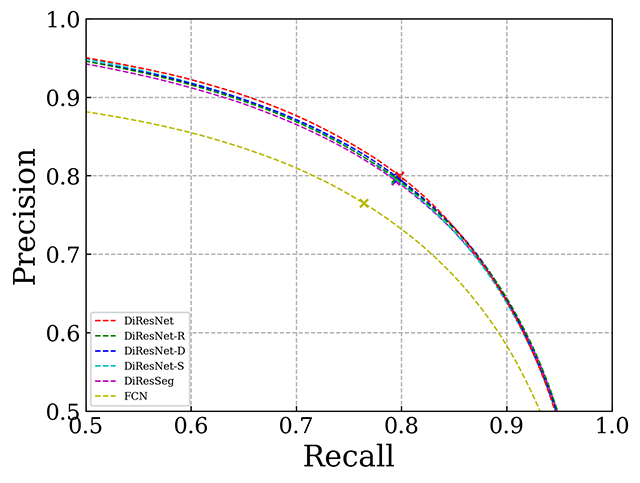}}
        \subcaptionbox{}
        {\includegraphics[width=4.3cm]{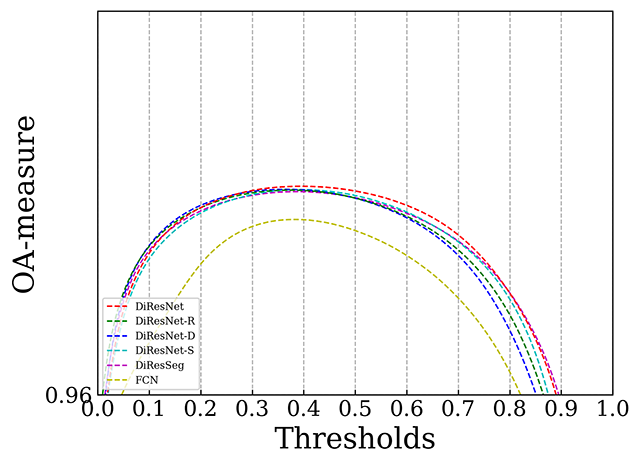}}
    \caption{Accuracy curves of the ablation study (Massachusetts dataset). (a) Precision-recall curves, (b) OA curves.}\label{Fig.AblationCurveMas}
\end{figure}

Table \ref{Table.AblationMas} reports the quantitative results of the ablation study on the Massachusetts dataset. DiResNet-R, DiResNet-S and DiResNet-D refer to the different versions of DiResNet with the DiResRef, structural supervision and direction supervision, respectively. Compared to the baseline FCN, DiResSeg shows a significant improvements in both OA and F1 measures. The use of DiResRef in DiResNet-R improves the results of 0.16\% in the F1 measure. Compared to the DiResSeg with only the segmentation network, DiResNet-S and DiResNet-D have advantages of 0.08\% and 0.23\% in the F1 measure, respectively. The DiResNet with all auxiliary designs shows an increase of 0.45\% in F1 measure and 0.08\% in OA. Compared with the baseline FCN, the DiResNet has an advantage of 0.43\% in OA and 3.35\% in F1 measure. Figure \ref{Fig.AblationCurveMas} shows the accuracy curves of the ablation study in the Massachusetts dataset. Figure \ref{Fig.AblationCurveMas}(a) shows the precision-recall curve, while Figure \ref{Fig.AblationCurveMas}(b) shows the calculated OA under different thresholds. One can observe that the designed DiResNet has a great advantage compared to the baseline FCN. The red curve represents the DiResNet, which has the biggest areas in both graphs.

\begin{table}[htbp]
    \centering
    \caption{Results of the ablation study (DeepGlobe dataset).}
    \resizebox{1\linewidth}{!}{%
        \begin{tabular}{l|cc|cc|cccc}
        \toprule
            \multirow{2}*{Method} & \multicolumn{2}{c|}{Components} & \multicolumn{2}{c|}{Supervisions} & \multirow{2}*{OA(\%)}  & \multirow{2}*{P(\%)}  & \multirow{2}*{R(\%)}  & \multirow{2}*{F1(\%)} \\
            \cline{2-5}
            & DiResSeg & DiResRef & Structure & Direction  \\
            \hline
            FCN & &  &  &  & 97.95 & 69.95 & 81.05 & 73.83\\
            \hline
            DiResSeg & $\surd$ &  &  & & 98.32 & 75.77 & 83.61 & 78.35\\
            DiResNet-R & $\surd$ & $\surd$ & & & 98.35 & 76.25 & 83.49 & 78.56 \\
            DiResNet-S & $\surd$ &  & $\surd$ &  & 98.38 & 77.53 & 82.08 & 78.55\\
            DiResNet-D & $\surd$ & &  & $\surd$ & 98.36 & 76.29 & \textbf{83.62} & 78.69\\
            DiResNet & $\surd$ & $\surd$ & $\surd$ & $\surd$ & \textbf{98.44} & \textbf{78.76} & 81.46 & \textbf{79.09}\\
        \bottomrule
        \end{tabular}
    }
    \label{Table.AblationDG}
\end{table}

\begin{figure}[htbp]
\centering
        \subcaptionbox{}
        {\includegraphics[width=4.3cm]{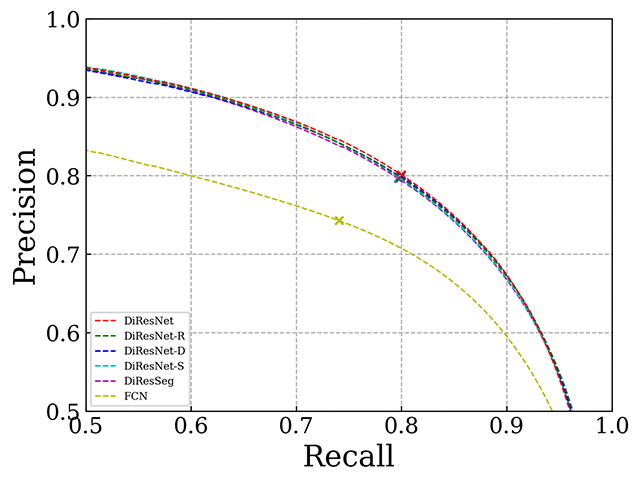}}
        \subcaptionbox{}
        {\includegraphics[width=4.3cm]{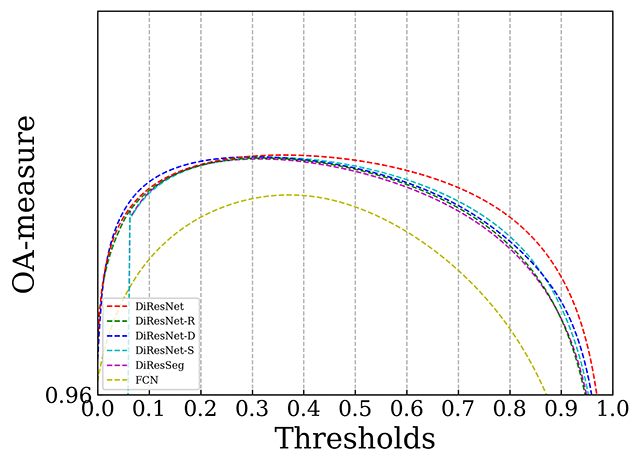}}
    \caption{Accuracy curves of the ablation study (DeepGlobe dataset). (a) Precision-recall curves, (b) OA curves.}\label{Fig.AblationCurveDG}
\end{figure}

Table \ref{Table.AblationDG} reports the quantitative results of the ablation study on the DeepGlobe dataset. Compared to the baseline FCN, the designed DiResSeg shows an improvement of 4.52\% and 0.37\% in F1 measure and OA, respectively. Compared to the DiResSeg, the DiResNet-R, DiResNet-S and DiResNet-D have advantages of 0.21\%, 0.20\% and 0.34\% in F1 measure, respectively. DiResNet shows an increase of 0.45\% in F1 measure and 0.06\% in OA compared to the DiResSeg. Its improvement over the baseline FCN is 3.35\% in F1 measure and 0.43\% in OA. The precision-recall and OA curves of the compared methods are reported in Fig.\ref{Fig.AblationCurveDG}. The DiResNet and its variations achieve significant advantages over the baseline FCN.

Fig.\ref{Fig.AblationResults} shows the segmentation results on several testing areas. Compared with the baseline FCN, the segmentation maps of DiResSeg are smoothed and less fragmented due to its deconvolutional layers. However, there are still many false alarms and interruptions in the segmented road maps. With the use of auxiliary designs in DiResNet, the false alarms are reduced. Additionally, the connectivity of roads is greatly improved.

\subsection{Analysis of the Effect of DiResRef}

\begin{figure}[thpb]
\centering
    {\includegraphics[height=0.5cm]{ablation_pic/BNsegColorBar.png}}\\
    \setlength{\tabcolsep}{1.0pt}
    \begin{tabular}{>{\centering\arraybackslash}m{0.5cm}>{\centering\arraybackslash}m{2cm}>{\centering\arraybackslash}m{2cm}>{\centering\arraybackslash}m{2cm}>{\centering\arraybackslash}m{2cm}}
        (a) &
        \includegraphics[width=2cm]{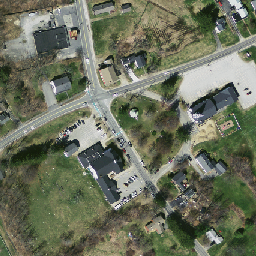} &
        \includegraphics[width=2cm]{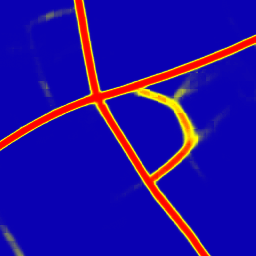} &
        \includegraphics[width=2cm]{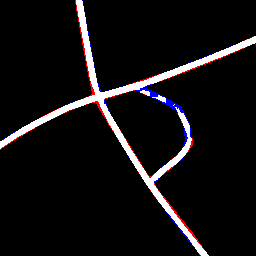} &
        \includegraphics[width=2cm]{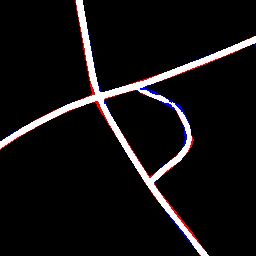}\\
        (b) &
        \includegraphics[width=2cm]{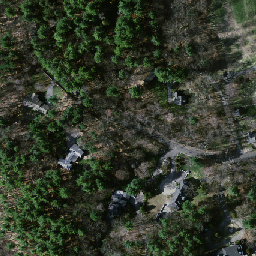} &
        \includegraphics[width=2cm]{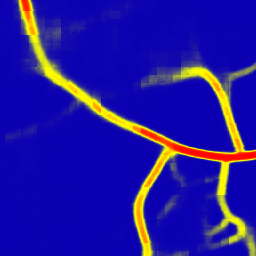} &
        \includegraphics[width=2cm]{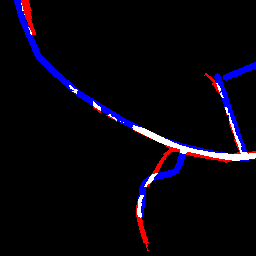} &
        \includegraphics[width=2cm]{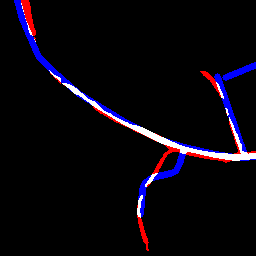}\\
        (c) &
        \includegraphics[width=2cm]{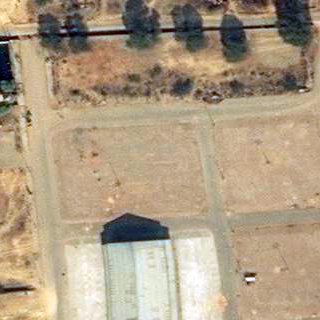} &
        \includegraphics[width=2cm]{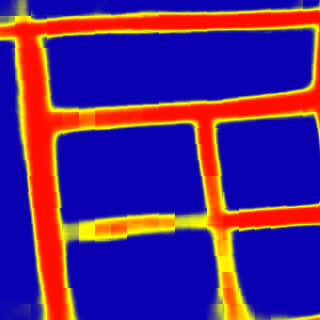} &
        \includegraphics[width=2cm]{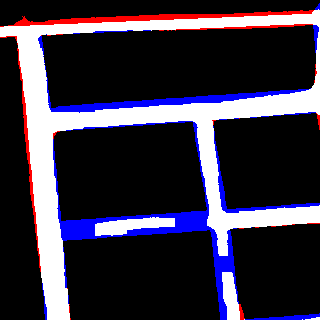} &
        \includegraphics[width=2cm]{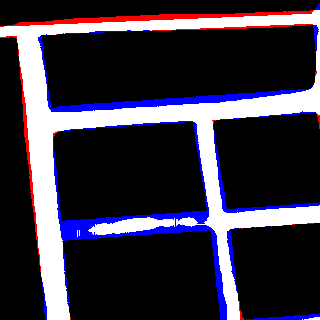}\\
        (d) &
        \includegraphics[width=2cm]{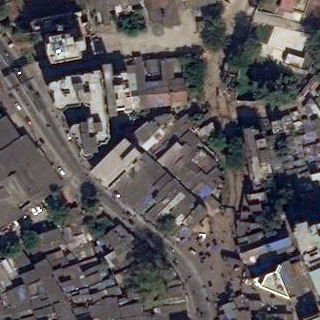} &
        \includegraphics[width=2cm]{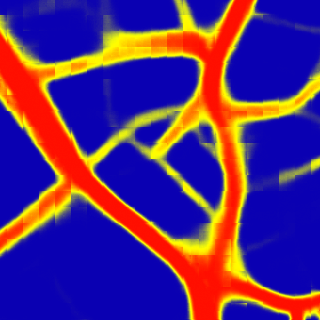} &
        \includegraphics[width=2cm]{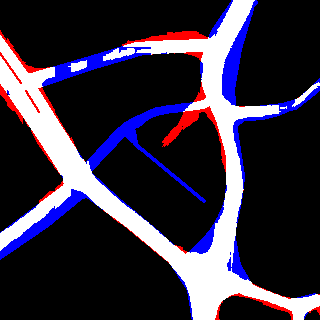} &
        \includegraphics[width=2cm]{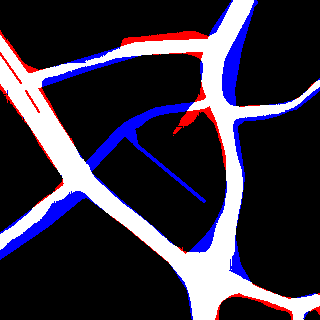}\\
        & Test image & Probability Map & Before DiResRef & After DiResRef\\
    \end{tabular}
    \caption{Examples illustrating the effect of DiResRef on sample testing areas. (a), (b) Results selected from the Massachusetts dataset, (c), (d) Results selected from the DeepGlobe dataset.}
    \label{Fig.RefineEffect}
\end{figure}

As presented in Table \ref{Table.AblationMas}, the DiResRef increases F1 measure of DiResSeg by 0.16\% in the Massachusetts dataset and by 0.21\% in the DeepGlobe dataset. To qualitatively evaluate the DiResRef, in Fig.\ref{Fig.RefineEffect} we compare the obtained results before and after its use. The optimization of DiResRef is based on the road probability maps produced by the network. Fig.\ref{Fig.RefineEffect}(a) and (b) show two sample areas in the Massachusetts dataset affected by occlusions (caused by trees). In the original road maps produced by the DiResSeg, there are interruptions on the roads. After using the DiResRef, some of the interrupted segments have been connected and the results are more complete. Fig.\ref{Fig.RefineEffect}(c) and (d) show two cases of wide roads in the DeepGlobe dataset. One can observe that the use of DiResRef not only connects some of the interruptions, but also smooths the road boundaries.

\subsection{Analysis of the Effect of Direction Supervision}
\begin{figure}[thpb]
\centering
    {\includegraphics[height=0.5cm]{ablation_pic/BNsegColorBar.png}}\\
    \setlength{\tabcolsep}{1.0pt}
    \begin{tabular}{>{\centering\arraybackslash}m{0.5cm}>{\centering\arraybackslash}m{2cm}>{\centering\arraybackslash}m{2cm}>{\centering\arraybackslash}m{2cm}>{\centering\arraybackslash}m{2cm}}
        (a) &
        \includegraphics[width=2cm]{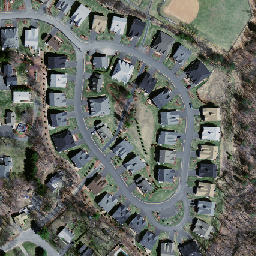} &
        \includegraphics[width=2cm]{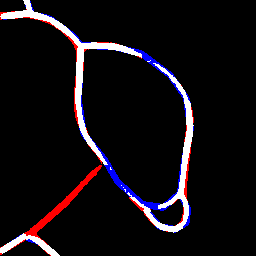} &
        \includegraphics[width=2cm]{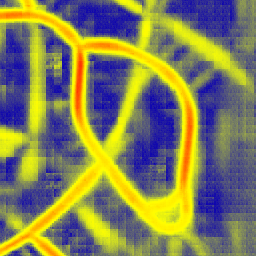} &
        \includegraphics[width=2cm]{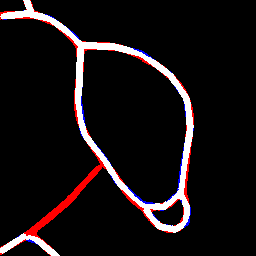}\\
        (b) &
        \includegraphics[width=2cm]{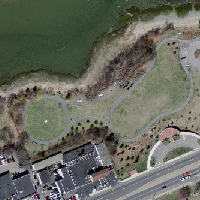} &
        \includegraphics[width=2cm]{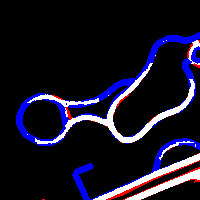} &
        \includegraphics[width=2cm]{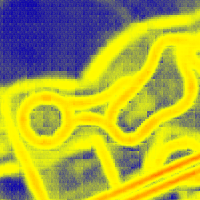} &
        \includegraphics[width=2cm]{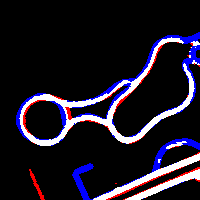}\\
        (c) &
        \includegraphics[width=2cm]{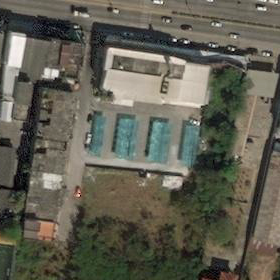} &
        \includegraphics[width=2cm]{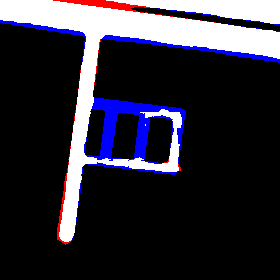} &
        \includegraphics[width=2cm]{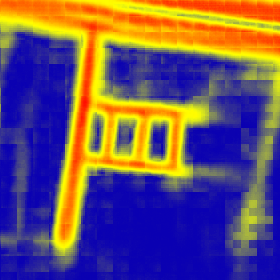} &
        \includegraphics[width=2cm]{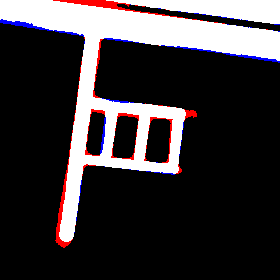}\\
        (d) &
        \includegraphics[width=2cm]{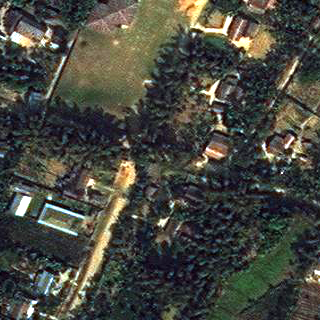} &
        \includegraphics[width=2cm]{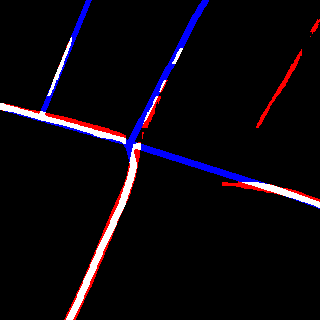} &
        \includegraphics[width=2cm]{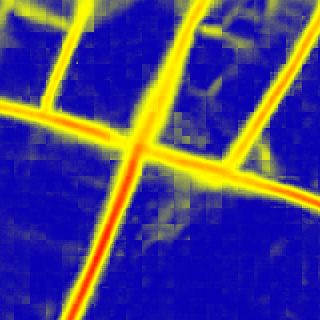} &
        \includegraphics[width=2cm]{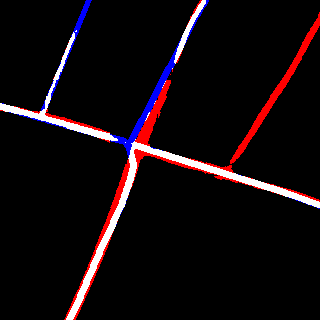}\\
        & Test image & Without Direction Supervison & Direction Salience Map & With Direction Supervision\\
    \end{tabular}
    \caption{Examples illustrating the effect of direction supervision on sample testing areas. (a), (b) Results selected from the Massachusetts dataset, (c), (d) Results selected from the DeepGlobe dataset.}
    \label{Fig.DirectionEffect}
\end{figure}

To visually assess the effect of direction supervision, in Fig.\ref{Fig.DirectionEffect} we compare the segmentation results with and without its use. The direction salience maps are generated by adding the 4 channel outputs of the predicted directions, which imply the linear features learned by the network. One can observe that the use of direction supervision enables the network to better embed the linear features, thus improving the detection of roads. As a result, the DiResNet-D with direction supervision obtains the best recall measure on both datasets. However, one of the side effects of the direction supervision is that it also enhances some of the non-road linear features.

\subsection{Comparative Experiments}
In this section we compare the proposed method with literature works. The compared methods include the baseline FCN \cite{he2016resnet}, the CasNet \cite{cheng2017casnet}, the original UNet \cite{ronneberger2015unet} and the residual UNet (ResUNet) \cite{zhang2018roadresunet}. CasNet is an early work on road segmentation using a modified version of VGG-Net \cite{simonyan2014vggnet} as encoder. To fairly compare the tested methods, the same data pre-processing procedures and parameter settings are used during the training. Table \ref{Table.CompareMas} reports the quantitative results of the compared methods on the Massachusetts dataset. The accuracy provided by the FCN is lower than those of other methods. This is mainly due to the sequential down-sampling operations in its early layers. The UNet has an advantage of 0.07\% in BEP compared with CasNet, but its computational cost is significantly higher. The residual design in ResUNet improves the F1 of 0.12\% and the BEP of 0.11\% compared with the original UNet. It also achieves the best recall measures.

\begin{figure}[htbp]
\centering
        \subcaptionbox{}
        {\includegraphics[width=4.3cm]{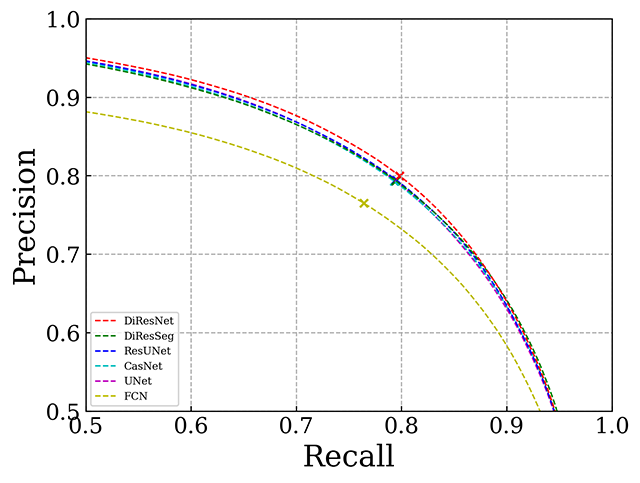}}
        \subcaptionbox{}
        {\includegraphics[width=4.3cm]{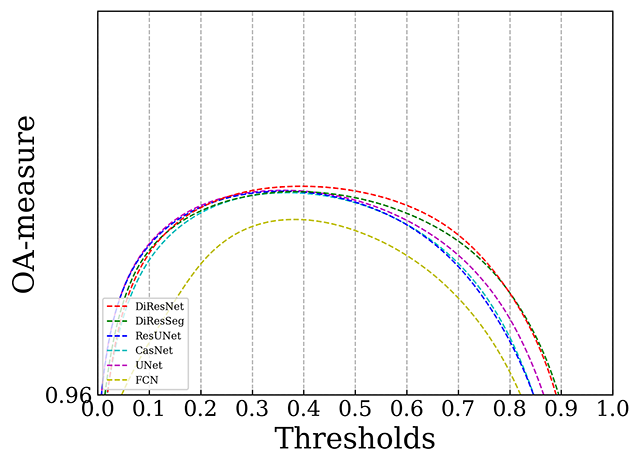}}
    \caption{Accuracy curves of the ablation study (Massachusetts dataset). (a) Precision-recall curves, (b) OA curves.}\label{Fig.CompareCurveMas}
\end{figure}

Compared with the literature works, the designed DiResSeg obtains the best results in terms of both F1 and OA. Considering that its computation cost is significantly lower than those of UNet and ResUNet, these improvements are remarkable. After adding the auxiliary supervisions and the DiResRef, the proposed DiResNet achieves the best performance on BEP, F1 and OA. In greater detail, it has an advantage of 0.60\% in F1 and 0.11\% in OA compared with UNet. The precision-recall and OA curves of the compared methods are presented in Fig.\ref{Fig.CompareCurveMas}. The OA of the proposed method is higher under all the thresholds.

\begin{table}[htbp]
    \centering
    \caption{Results of the comparative experiments (Massachusetts dataset).}
    \resizebox{1\linewidth}{!}{%
        \begin{tabular}{l|c|cc|cc}
        \toprule
            Method & OA(\%) & P(\%)  & R(\%) & BEP(\%) & F1(\%) \\
            \hline
            FCN \cite{he2016resnet} &97.70 & 74.25 & 78.93 & 76.46 & 76.35 \\
            CasNet \cite{cheng2017casnet} & 97.99 & 77.65 & 80.87 & 79.29 & 79.06 \\
            UNet \cite{ronneberger2015unet} & 98.02 & 78.20 & 80.46 & 79.36 & 79.10 \\
            ResUNet \cite{zhang2018roadresunet} & 98.00 & 77.69 & \textbf{81.14} & 79.47 & 79.23 \\
            \hline
            DiResSeg & 98.07 & 79.44 & 79.55 & 79.40 & 79.33 \\
            DiResNet & \textbf{98.13} & \textbf{80.38} & 79.41 & \textbf{79.90} & \textbf{79.70}\\
        \bottomrule
        \end{tabular}
    }
    \label{Table.CompareMas}
\end{table}

\begin{figure}[htbp]
\centering
        \subcaptionbox{}
        {\includegraphics[width=4.3cm]{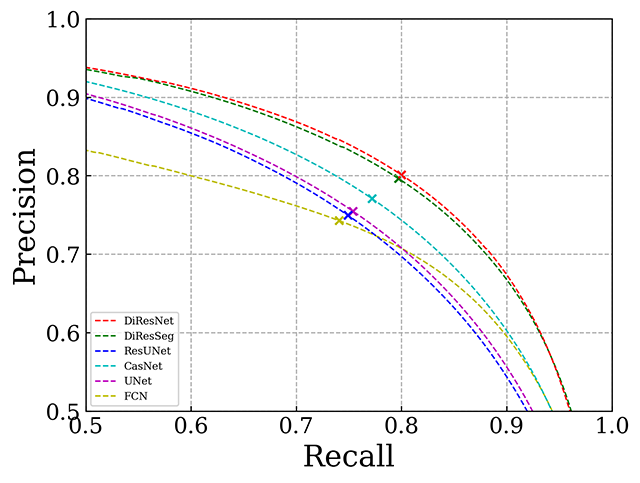}}
        \subcaptionbox{}
        {\includegraphics[width=4.3cm]{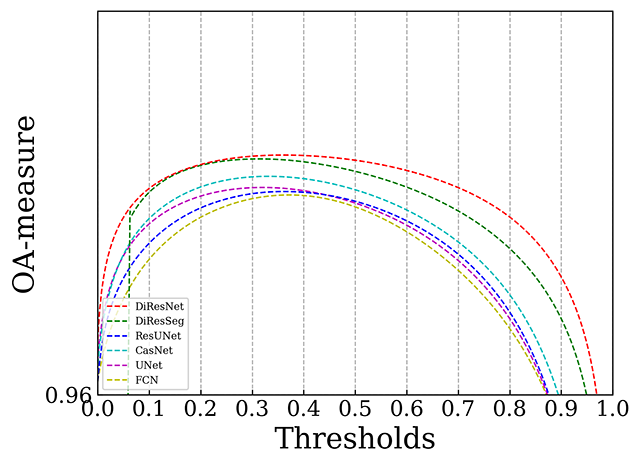}}
    \caption{Accuracy curves of the ablation study (DeepGlobe dataset). (a) Precision-recall curves, (b) OA curves.}\label{Fig.CompareCurveDG}
\end{figure}

Table \ref{Table.CompareMas} reports the quantitative results obtained on the DeepGlobe dataset. This dataset is more challenging since there are more types of road surfaces included. As a results, the UNet-like architectures (UNet and ResUNet) show disadvantages compared with other methods that use powerful encoder networks. Compared with UNet, the CasNet with VGG-Net as its encoder has a advantage of 1.78\% in F1 measure and 0.14\% in OA. The proposed DiResSeg and DiResNet obtain significant improvements in all the measures (especially OA, BEP and F1 measures). Particularly, the DiResNet has a great improvement in terms of precision. We attribute these improvements to two factors: i) The DeepGlobe dataset has a small GSD, thus the low-level concatenation designs in UNet-like architectures propagate more noise, which degrades the accuracy; ii) The auxiliary designs in DiResNet enhance the embedding of both road typologies and linear features, which greatly improves the precision. Fig.\ref{Fig.CompareCurveDG} presents different accuracy curves of the compared methods. The curves of DiResSeg and DiResNet cover the biggest areas, proving the highest accuracy performance of the proposed method.

\begin{table}[htbp]
    \centering
    \caption{Results of the comparative experiments (DeepGlobe dataset).}
    \resizebox{1\linewidth}{!}{%
        \begin{tabular}{l|c|cc|cc}
        \toprule
            Method & OA(\%) & P(\%)  & R(\%) & BEP(\%) & F1(\%) \\
            \hline
            FCN \cite{he2016resnet} & 97.95 & 69.95 & 81.05 & 74.17 & 73.83 \\
            CasNet \cite{cheng2017casnet} & 98.13 & 73.66 & 80.63 & 77.24 & 75.73 \\
            UNet \cite{ronneberger2015unet} & 97.99  & 71.82 & 79.07 & 75.43 & 73.95 \\
            ResUNet \cite{zhang2018roadresunet} & 98.01 & 73.69 & 76.24 & 74.93 & 73.49 \\
            \hline
            DiResSeg & 98.32 & 75.77 & \textbf{83.63} & 79.67 & 78.35 \\
            DiResNet & \textbf{98.44} & \textbf{78.76} & 81.46 & \textbf{80.06} & \textbf{79.09}\\
        \bottomrule
        \end{tabular}
    }
    \label{Table.CompareDG}
\end{table}

\begin{figure*}[thpb]
\centering
    {\includegraphics[height=0.5cm]{ablation_pic/BNsegColorBar.png}}\\
    \setlength{\tabcolsep}{1pt}
    \begin{tabular}{>{\centering\arraybackslash}m{0.5cm}>{\centering\arraybackslash}m{3.5cm}>{\centering\arraybackslash}m{3.5cm}>{\centering\arraybackslash}m{3.5cm}>{\centering\arraybackslash}m{3.5cm}>{\centering\arraybackslash}m{3.5cm}}
        (a)&
        \includegraphics[width=3.5cm]{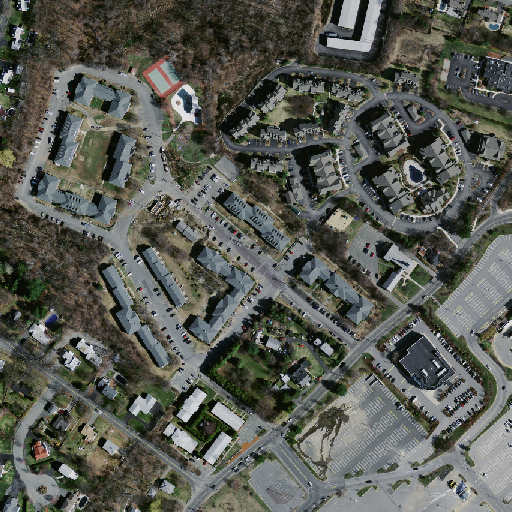} &
        \includegraphics[width=3.5cm]{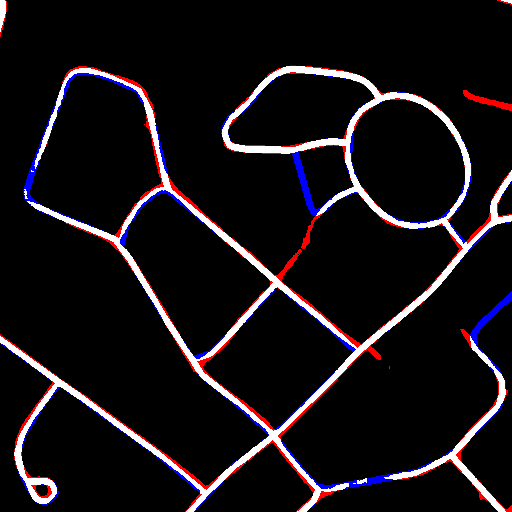} &
        \includegraphics[width=3.5cm]{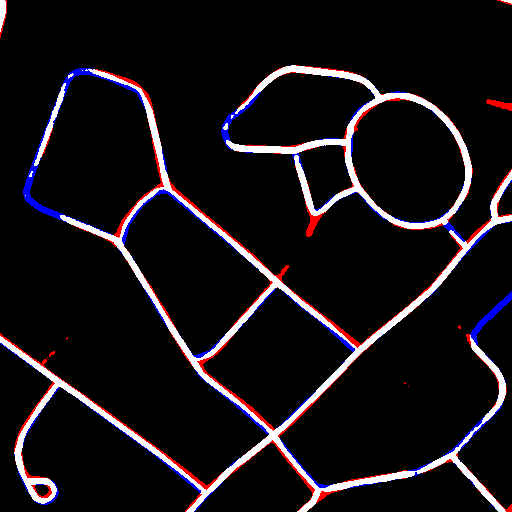} &
        \includegraphics[width=3.5cm]{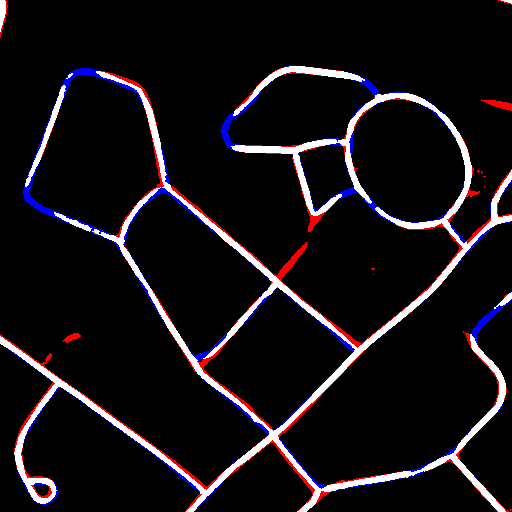} &
        \includegraphics[width=3.5cm]{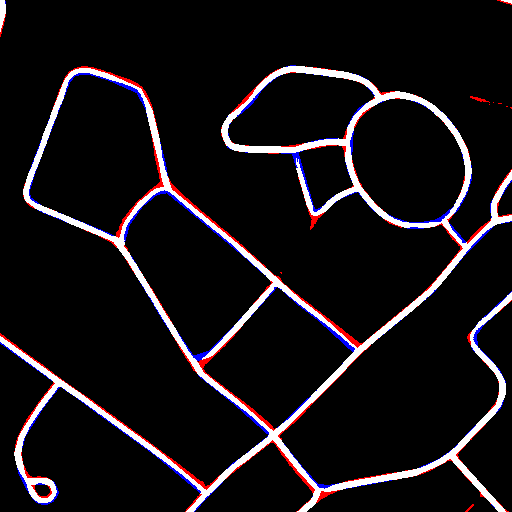}\\
        (b)&
        \includegraphics[width=3.5cm]{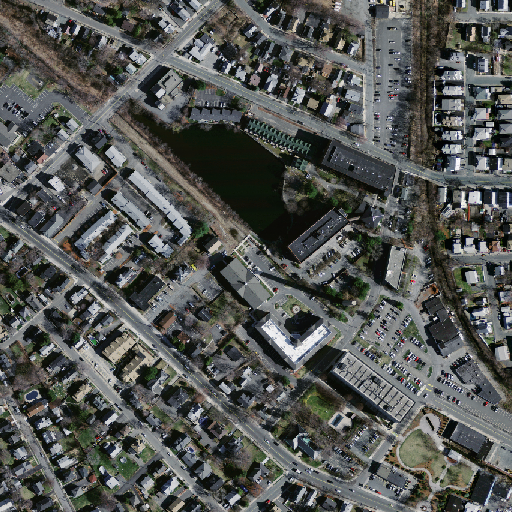} &
        \includegraphics[width=3.5cm]{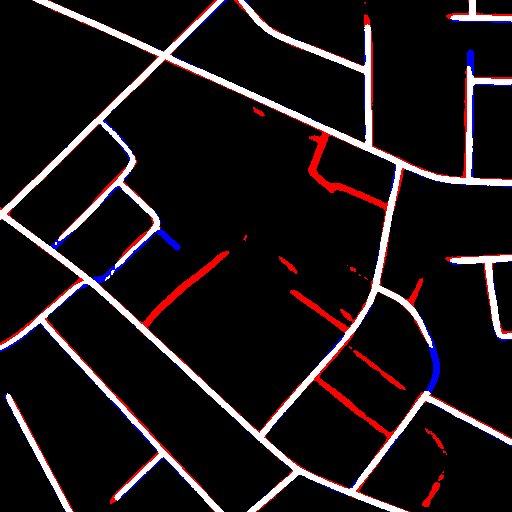} &
        \includegraphics[width=3.5cm]{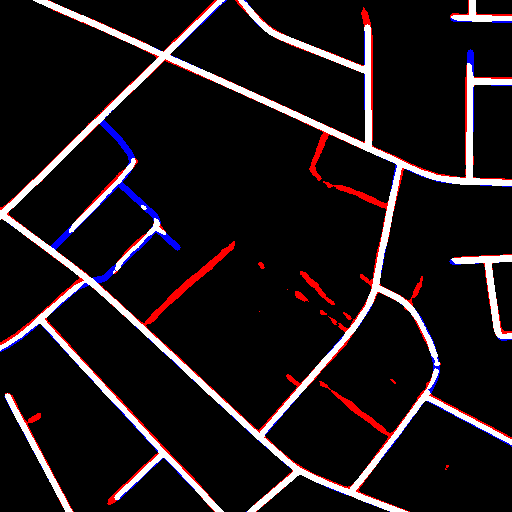} &
        \includegraphics[width=3.5cm]{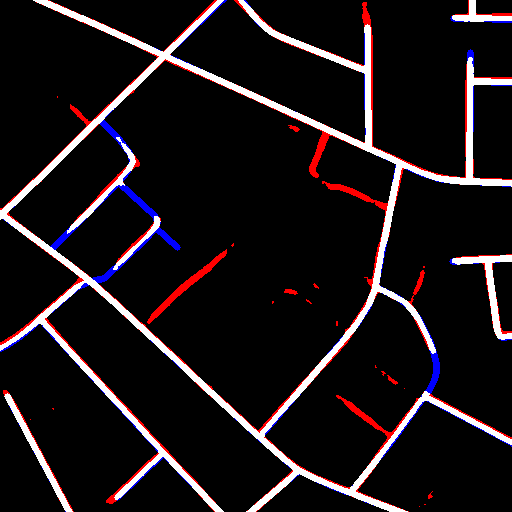} &
        \includegraphics[width=3.5cm]{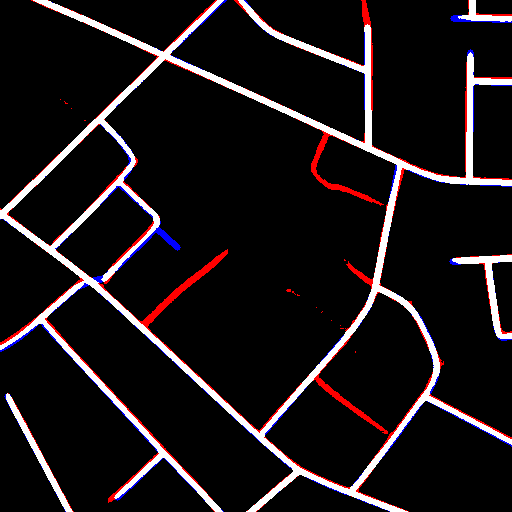}\\
        (c)&
        \includegraphics[width=3.5cm]{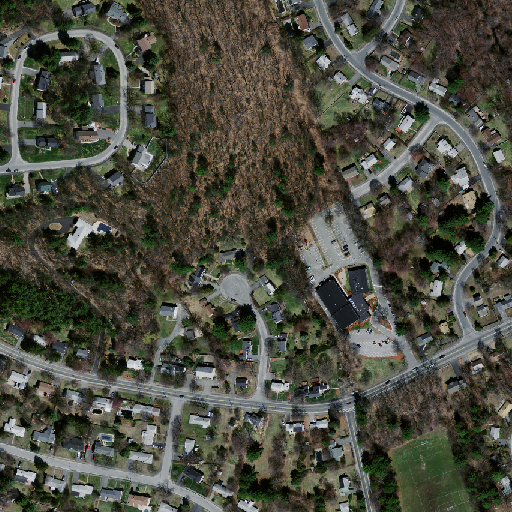} &
        \includegraphics[width=3.5cm]{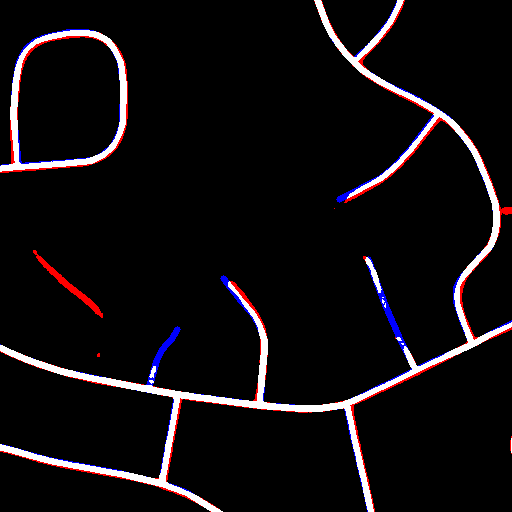} &
        \includegraphics[width=3.5cm]{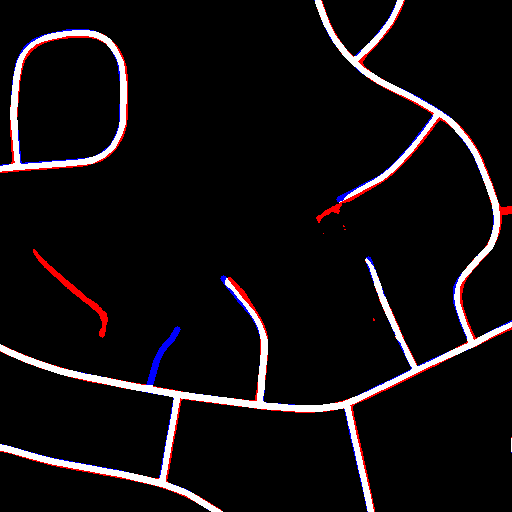} &
        \includegraphics[width=3.5cm]{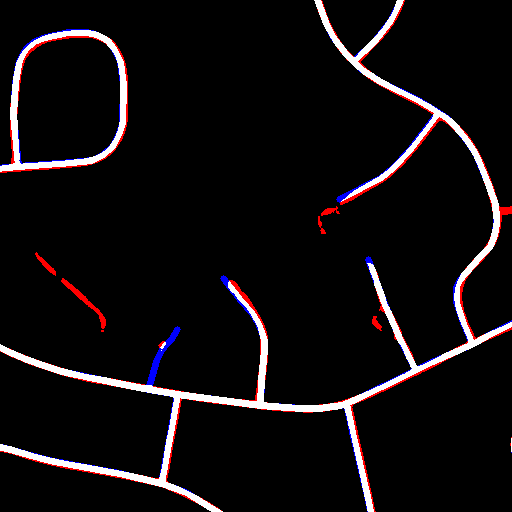} &
        \includegraphics[width=3.5cm]{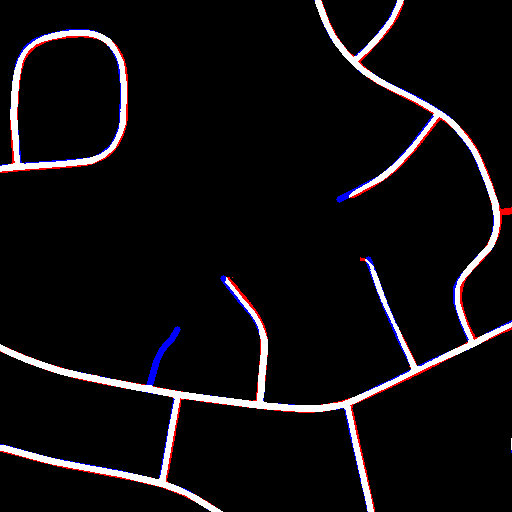}\\
        (d)&
        \includegraphics[width=3.5cm]{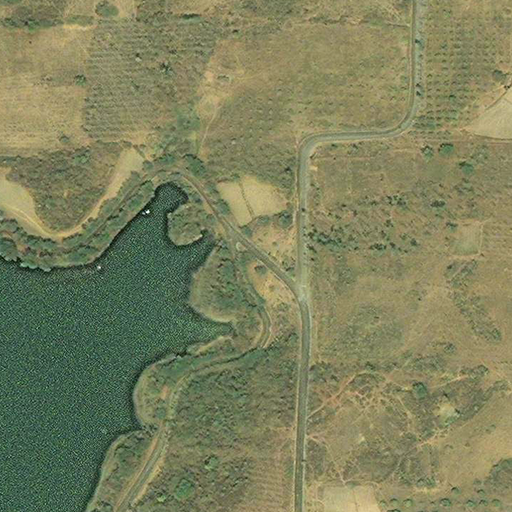} &
        \includegraphics[width=3.5cm]{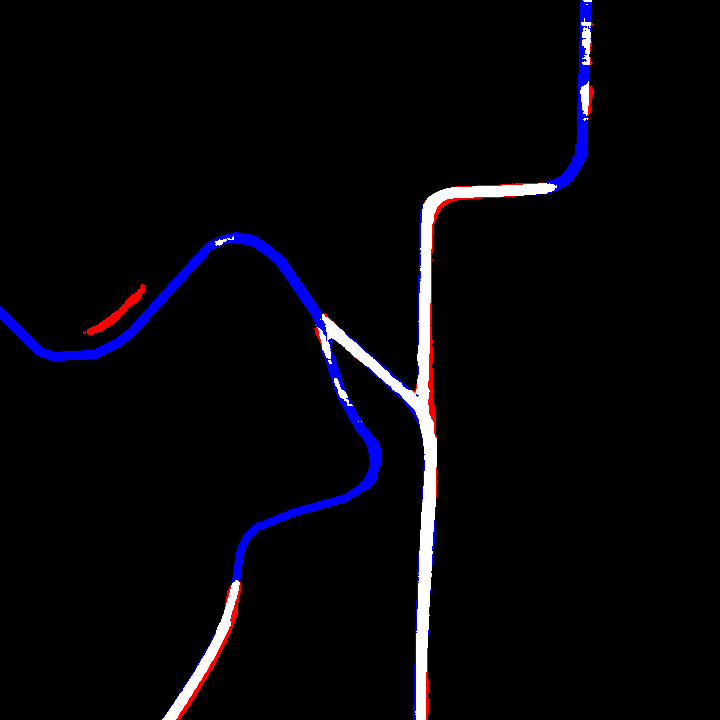} &
        \includegraphics[width=3.5cm]{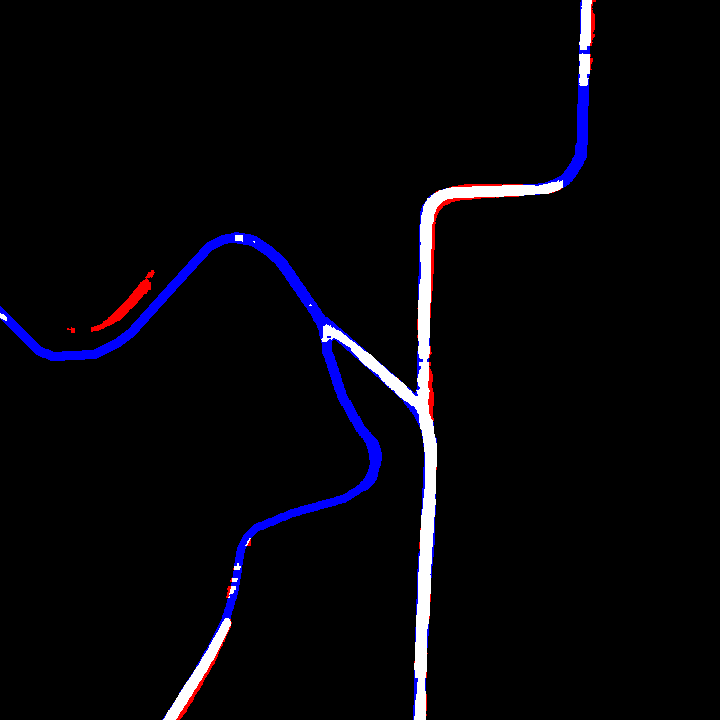} &
        \includegraphics[width=3.5cm]{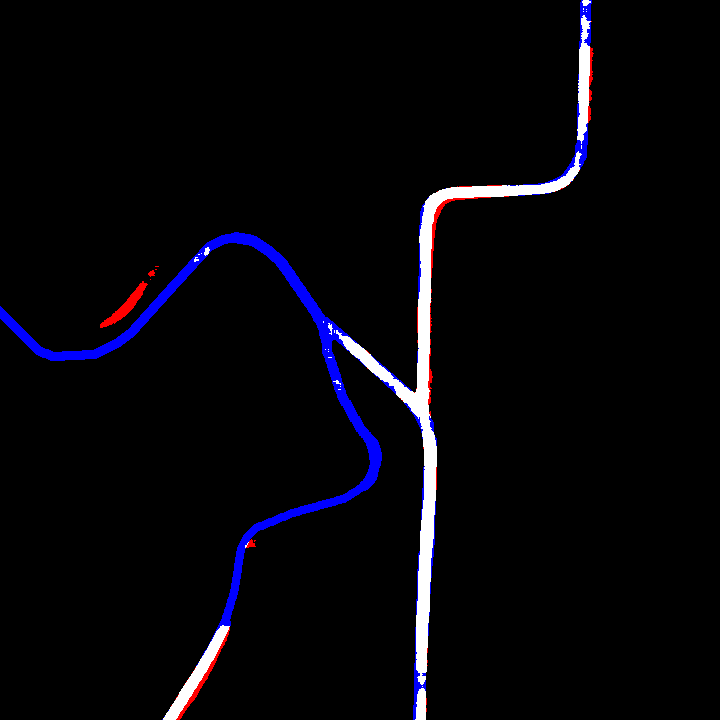} &
        \includegraphics[width=3.5cm]{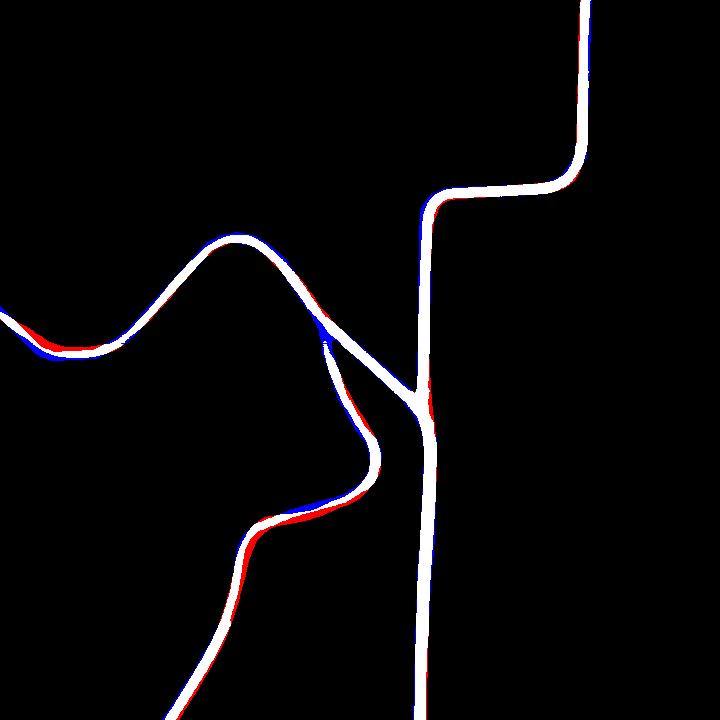}\\
        (e)&
        \includegraphics[width=3.5cm]{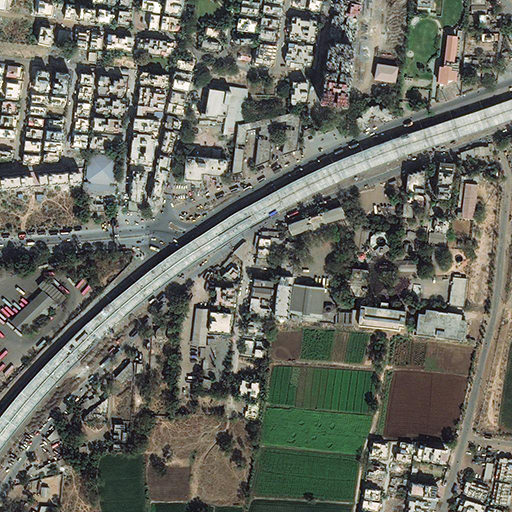} &
        \includegraphics[width=3.5cm]{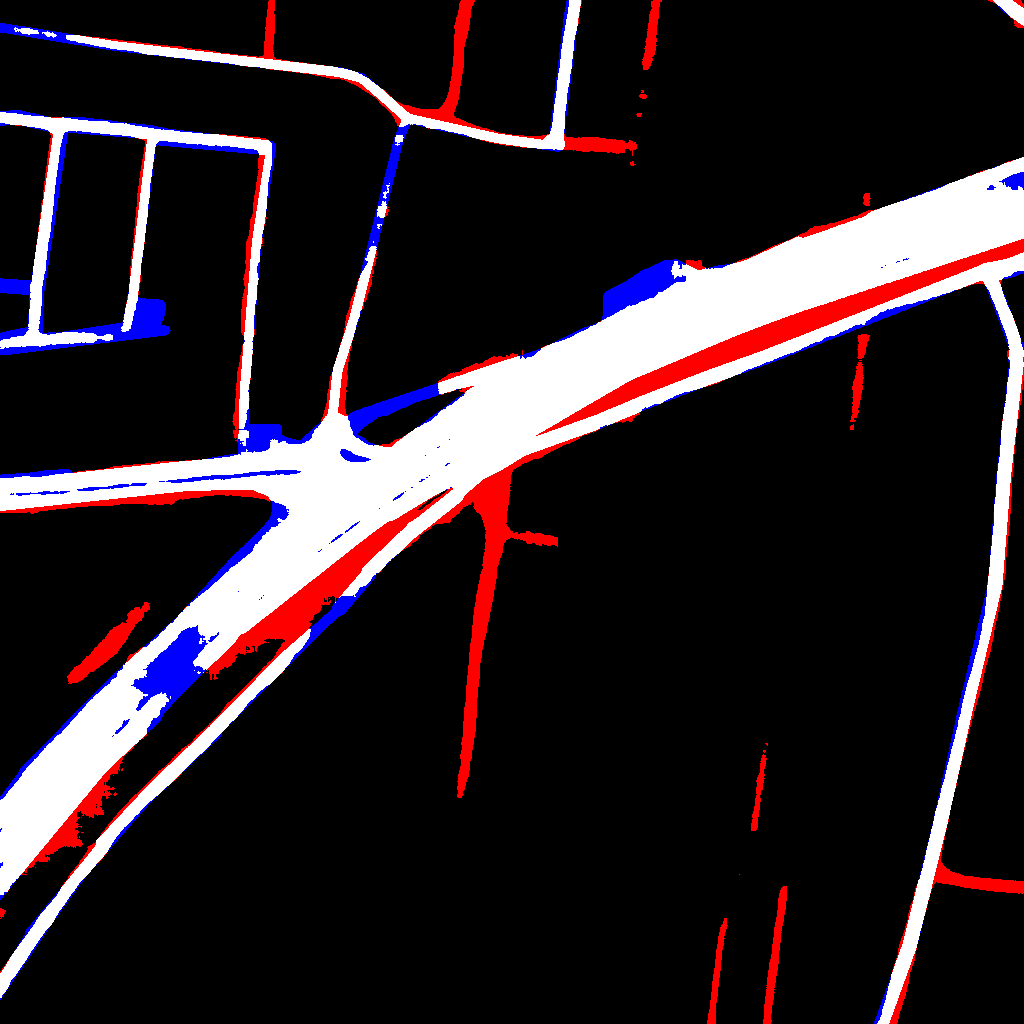} &
        \includegraphics[width=3.5cm]{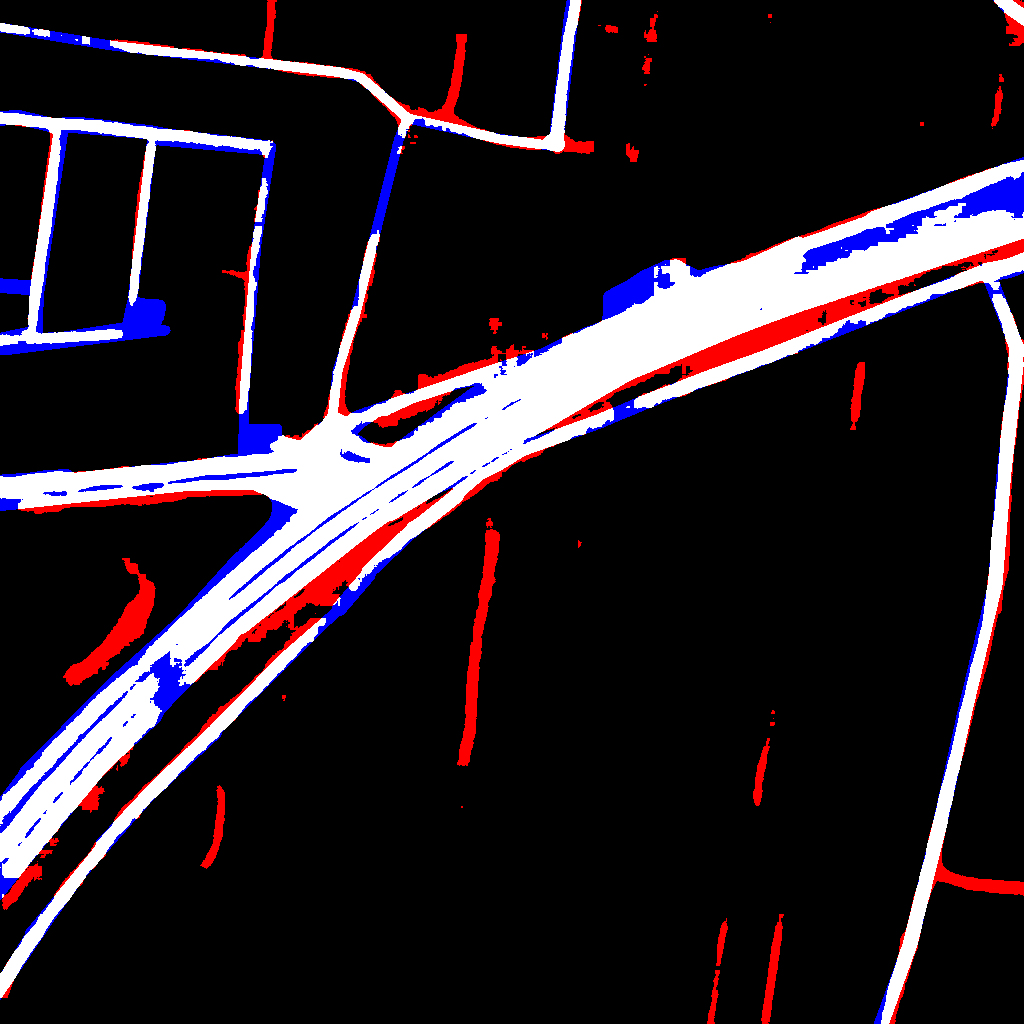} &
        \includegraphics[width=3.5cm]{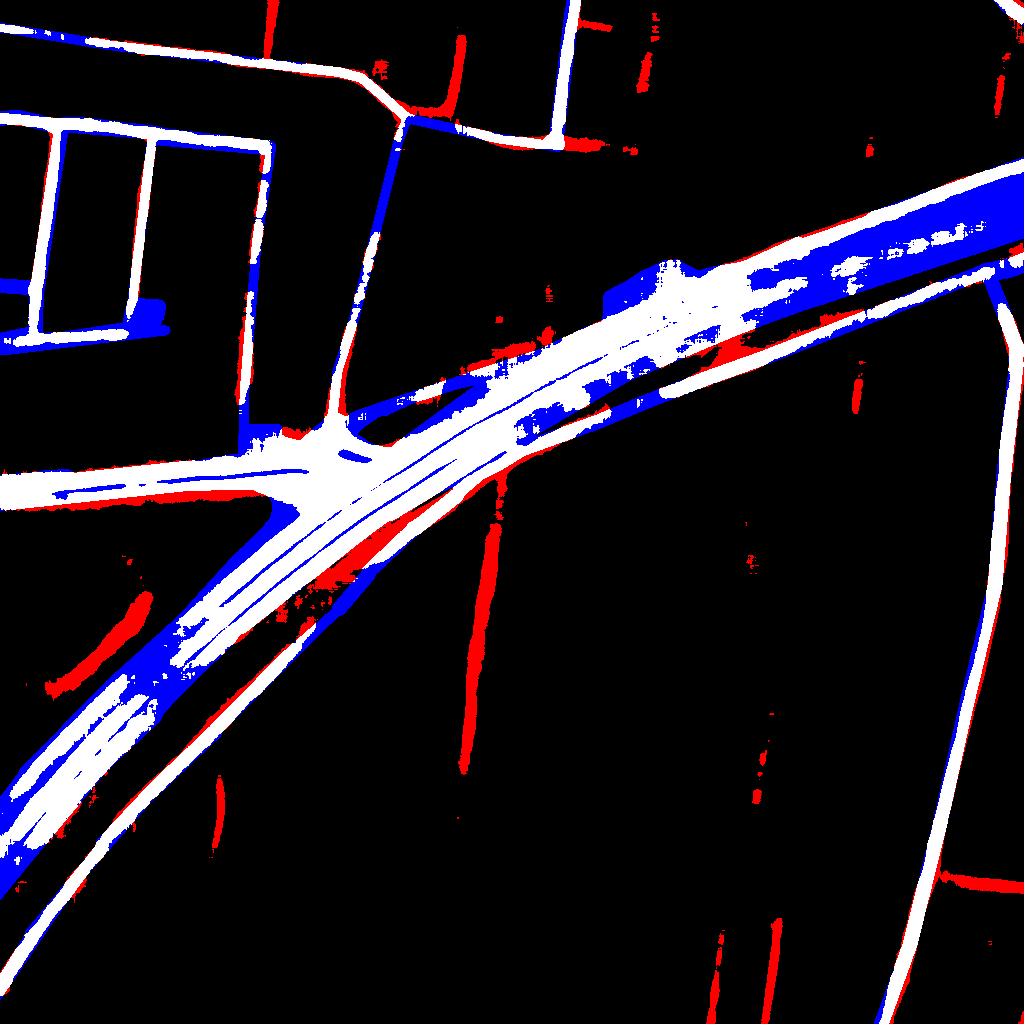} &
        \includegraphics[width=3.5cm]{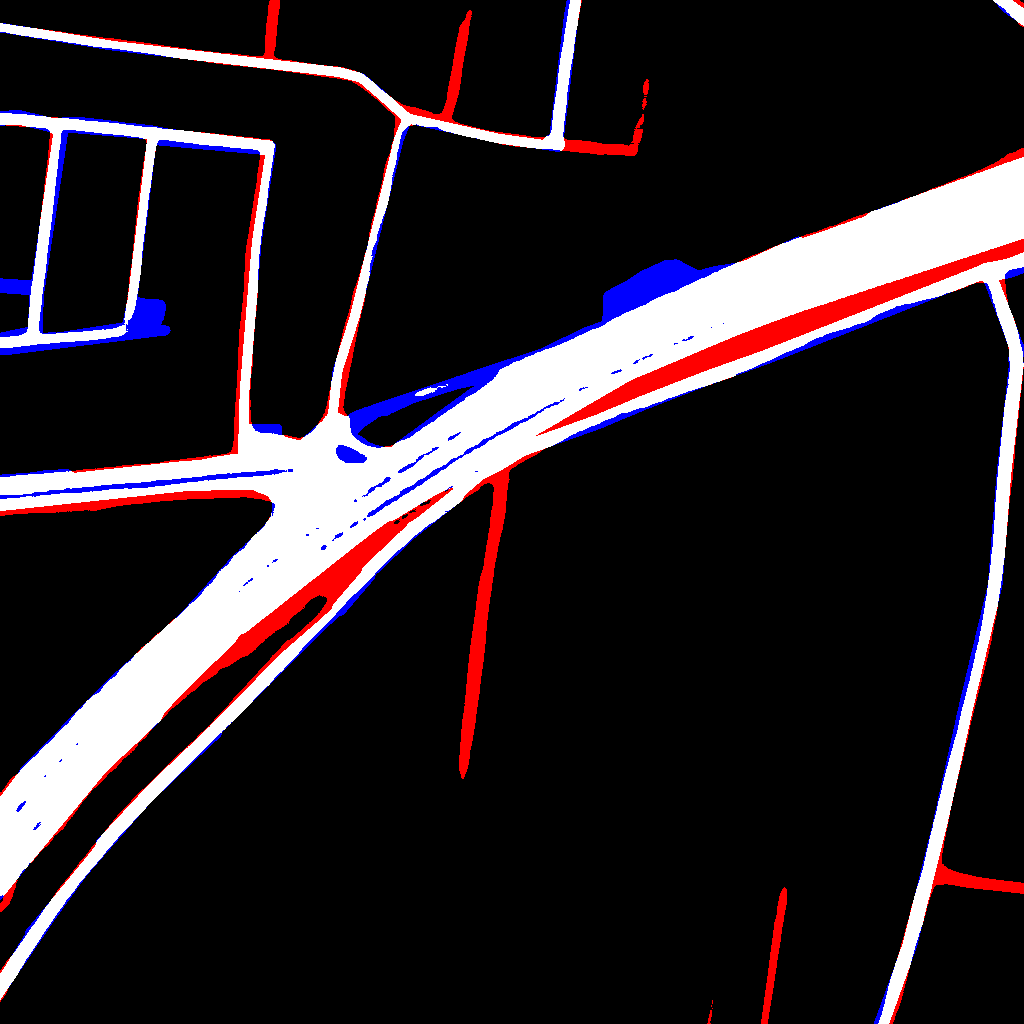}\\
        (f)&
        \includegraphics[width=3.5cm]{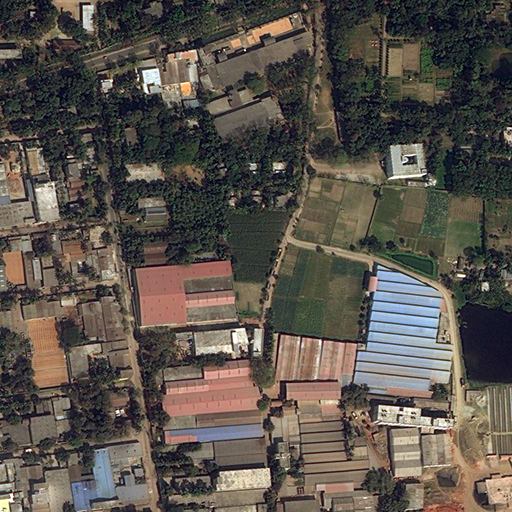} &
        \includegraphics[width=3.5cm]{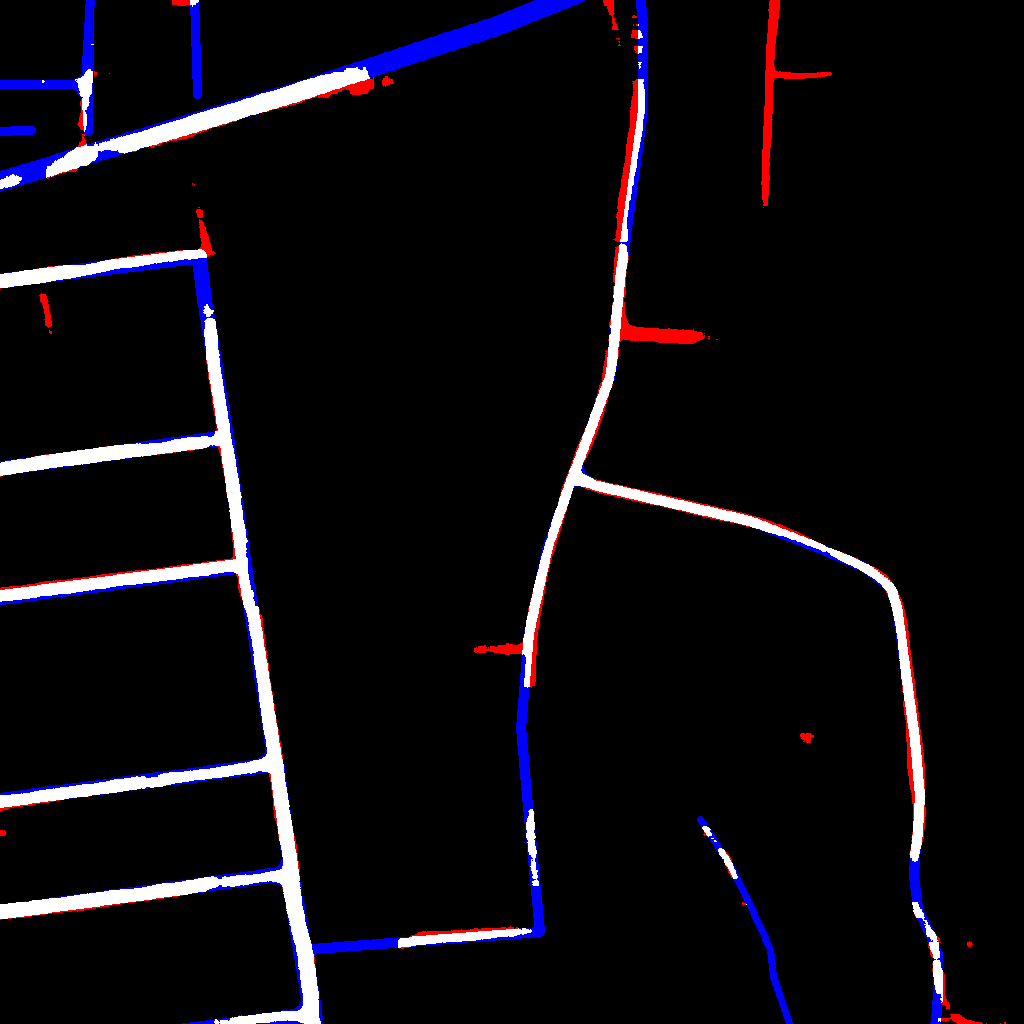} &
        \includegraphics[width=3.5cm]{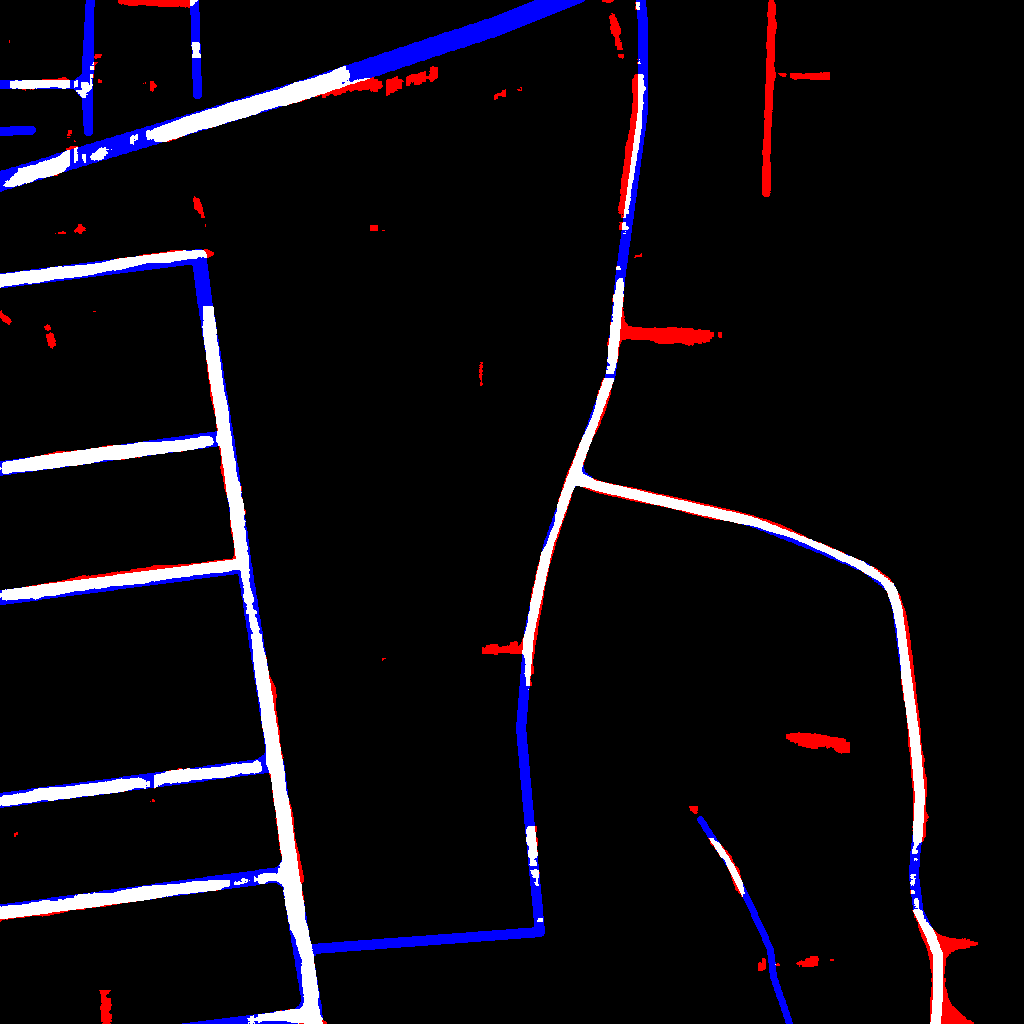} &
        \includegraphics[width=3.5cm]{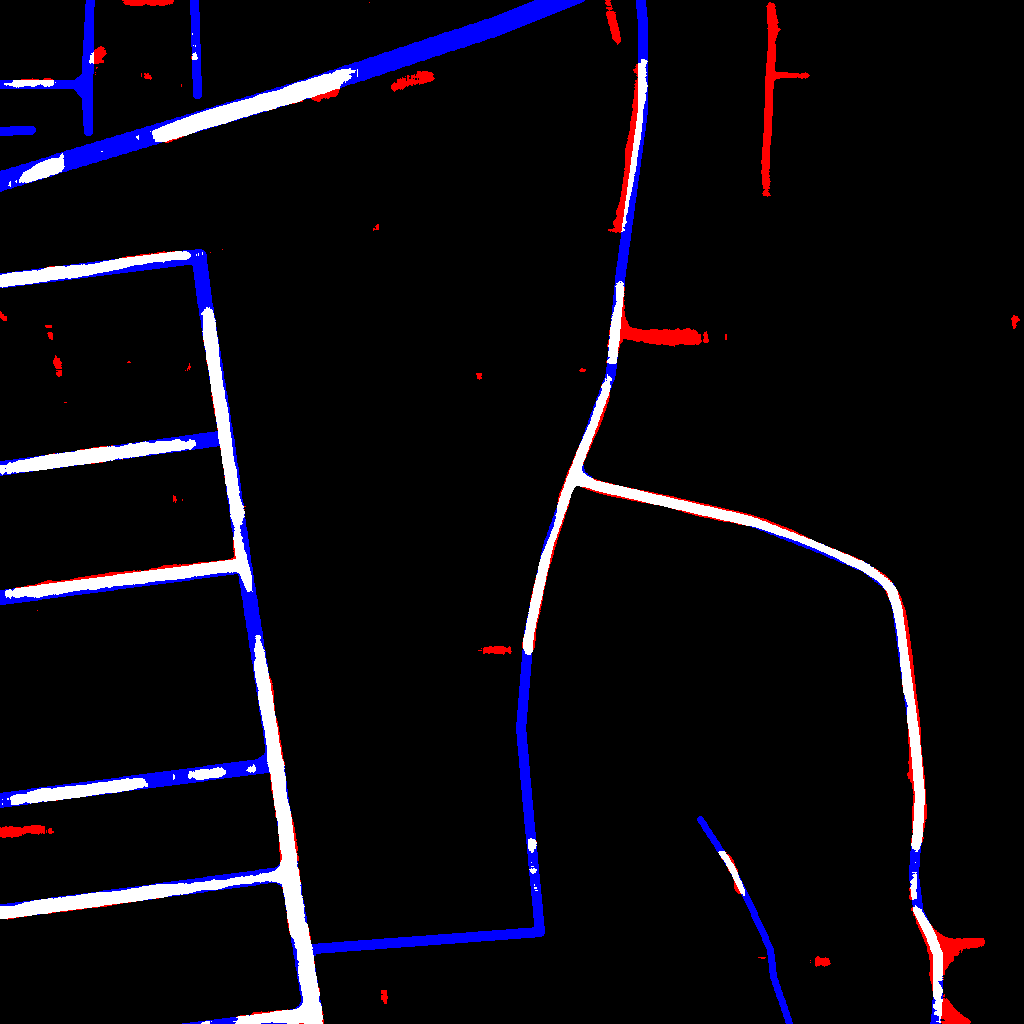} &
        \includegraphics[width=3.5cm]{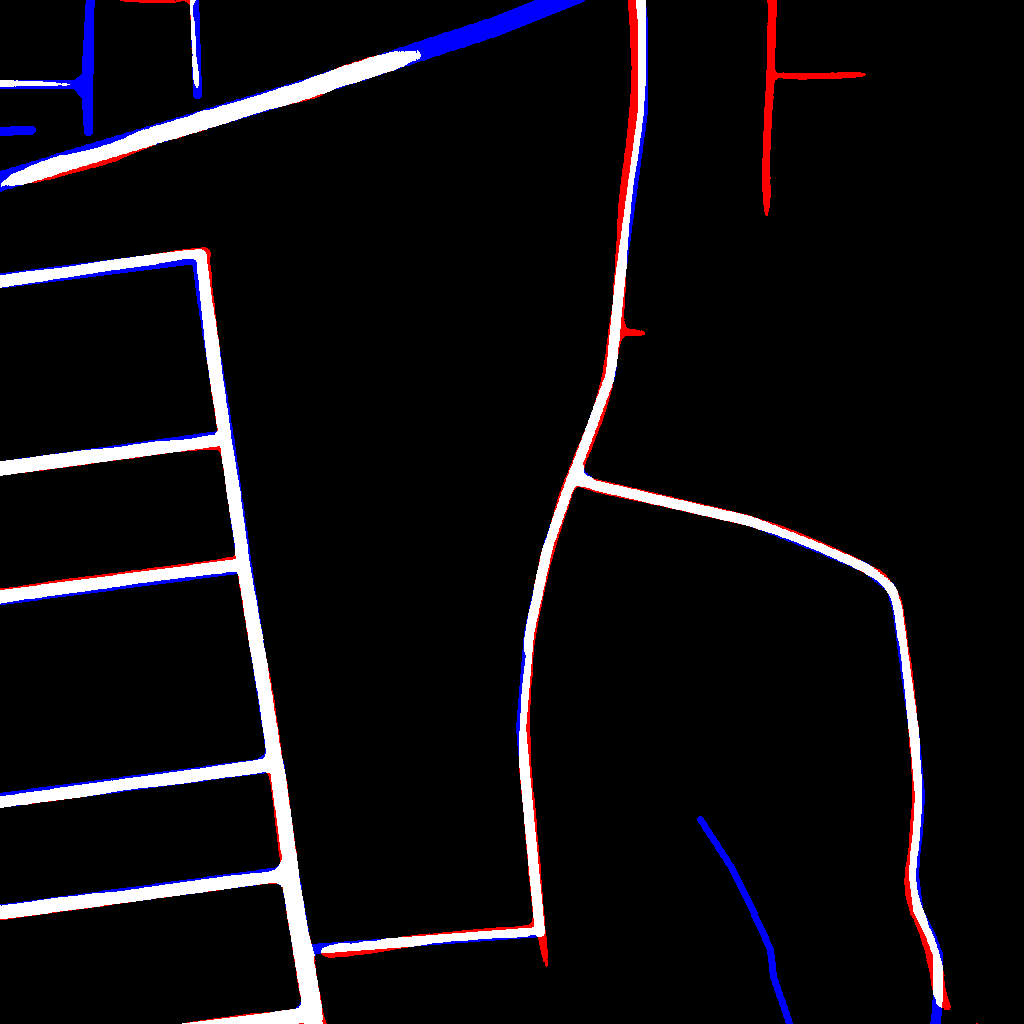}\\
        & Test image & CasNet & UNet & ResUNet & DiResNet\\
    \end{tabular}
    \caption{Examples of the segmentation results produced by the proposed method and the other used methods in the comparison. (a)-(c) Results selected from the Massachusetts dataset, (d)-(f) Results selected from the DeepGlobe dataset.}
    \label{Fig.CompareResults}
\end{figure*}

Fig.\ref{Fig.CompareResults} presents examples qualitative of the results obtained by different methods. The results of UNet and ResUNet are quite close. They both contain many interruptions and false alarms. In the areas with wide roads (e.g. Fig.\ref{Fig.CompareResults}(e)), this disadvantage is more severe. The CasNet obtains smoother results on the DeepGlobe dataset (Fig.\ref{Fig.CompareResults}(d)-(f)), but with more false alarms than on the Massachusetts dataset (Fig.\ref{Fig.CompareResults}(d)-(f)). Compared with its competitors, the proposed DiResNet shows two major advantages: i) It produces less false alarms, as the auxiliary supervisions enhance the precision of the results. ii) It produces more complete and smooth results. Indeed, the DiResNet significantly reduces the broken segments and strengthens the linear features.

\section{Conclusions}\label{sc6}
In this paper we have studied the CNN-based extraction of roads in VHR RSIs and presented a direction-aware residual network (DiResNet). The literature works mostly use UNet-like symmetric architectures, which are time-consuming and pass through more noises. We have argued and shown that the concatenation with low-level features is unnecessary and introduced an asymmetric network. This network is an extension of the Resnet, where its encoding layers are re-arranged to suit the task of road extraction. Experimental results on two benchmark datasets (the Massachusetts dataset and the DeepGlobe dataset) show that the designed segmentation network (DiResSeg) outperforms competitors in OA and F1 measures, whereas its computational cost is significantly smaller.

Moreover, we have presented three auxiliary designs to improve the road segmentation accuracy: i) a structure supervision to emphasize the preservation of the road topologies, ii) a direction supervision, where the angular operators are used to generate reference direction maps, thus supervising the network to learn directional linear features, and iii) a refinement sub-net to improve the smoothness and connectivity of the generated road maps. Combining these designs, the proposed DiResNet obtains sharp improvements in OA, BEP and F1 measures. It is worth noting that the precision of the proposed method is particularly high.

One of the remaining problems in road extraction is that there are a certain types of road surfaces that are neglected by all the considered network models. In the future, we will investigate to add nodes and length based analysis of the road networks to improve the generalization ability of the segmentation network.

\bibliographystyle{IEEEtran}
\bibliography{refs}

\begin{IEEEbiography}[{\includegraphics[width=1in,height=1.25in,clip,keepaspectratio]{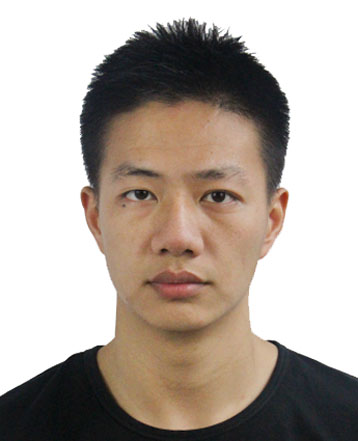}}]{Lei Ding}
received the B.S.degree in Measurement and Control Engineering in 2013 and the M.S. degree in Photogrammetry and Remote Sensing in 2016, both from University of Information Engineering, Zhengzhou, China. He is currently pursuing the Ph.D. degree at RSLab in the Department of Information Engineering and Computer Science, University of Trento, Italy. His research interests are related to remote sensing image processing and machine learning.
\end{IEEEbiography}

\begin{IEEEbiography}[{\includegraphics[width=1in,height=1.25in,clip,keepaspectratio]{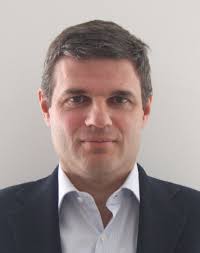}}]{Lorenzo Bruzzone}
(S'95-M'98-SM'03-F'10) received the Laurea (M.S.) degree in electronic engineering (\emph{summa cum laude}) and the Ph.D. degree in telecommunications from the University of Genoa, Italy, in 1993 and 1998, respectively. \\
He is currently a Full Professor of telecommunications at the University of Trento, Italy, where he teaches remote sensing, radar, and digital communications. Dr. Bruzzone is the founder and the director of the Remote Sensing Laboratory in the Department of Information Engineering and Computer Science, University of Trento. His current research interests are in the areas of remote sensing, radar and SAR, signal processing, machine learning and pattern recognition. He promotes and supervises research on these topics within the frameworks of many national and international projects. He is the Principal Investigator of many research projects. Among the others, he is the Principal Investigator of the \emph{Radar for icy Moon exploration} (RIME) instrument in the framework of the \emph{JUpiter ICy moons Explorer} (JUICE) mission of the European Space Agency. He is the author (or coauthor) of 215 scientific publications in referred international journals (154 in IEEE journals), more than 290 papers in conference proceedings, and 21 book chapters. He is editor/co-editor of 18 books/conference proceedings and 1 scientific book. He was invited as keynote speaker in more than 30 international conferences and workshops. Since 2009 he is a member of the Administrative Committee of the IEEE Geoscience and Remote Sensing Society (GRSS). 

Dr. Bruzzone ranked first place in the Student Prize Paper Competition of the 1998 IEEE International Geoscience and Remote Sensing Symposium (IGARSS), Seattle, July 1998. Since that he was recipient of many international and national honors and awards, including the recent IEEE GRSS 2015 Outstanding Service Award and the 2017 IEEE IGARSS Symposium Prize Paper Award. Dr. Bruzzone was a Guest Co-Editor of many Special Issues of international journals. He is the co-founder of the IEEE International Workshop on the Analysis of Multi-Temporal Remote-Sensing Images (MultiTemp) series and is currently a member of the Permanent Steering Committee of this series of workshops. Since 2003 he has been the Chair of the SPIE Conference on Image and Signal Processing for Remote Sensing. He has been the founder of the IEEE Geoscience and Remote Sensing Magazine for which he has been Editor-in-Chief between 2013-2017. Currently he is an Associate Editor for the IEEE Transactions on Geoscience and Remote Sensing. He has been Distinguished Speaker of the IEEE Geoscience and Remote Sensing Society between 2012-2016. His papers are highly cited, as proven form the total number of citations (more than 27000) and the value of the h-index (78) (source: Google Scholar).
\end{IEEEbiography}

\end{document}